%% file: main.tex
\documentclass[11pt]{article}

\usepackage[preprint]{acl}
\providecommand{\shortauthors}{}

\usepackage{times}
\usepackage{latexsym}
\usepackage[T1]{fontenc}
\usepackage[utf8]{inputenc}
\usepackage{microtype}
\usepackage{inconsolata}
\usepackage{lineno}
\usepackage{graphicx}
\usepackage{tabularx}
\usepackage{cuted}
\usepackage[utf8]{inputenc} 
\usepackage[T1]{fontenc}    
\usepackage{longtable, array}
\usepackage{ltablex}
\usepackage{booktabs}
\usepackage{subcaption} 
\usepackage{url}
\usepackage{graphicx}
\usepackage{amssymb}
\usepackage{amsfonts}
\usepackage{url}
\usepackage{bbm}
\usepackage{longtable}
\usepackage{rotating}
\usepackage{multirow}
\usepackage{mathrsfs}
\usepackage{enumitem}
\usepackage{adjustbox}
\usepackage{hyperref}
\usepackage{pgfplots}
\usetikzlibrary{pgfplots.dateplot}
\usepackage{filecontents}
\usepackage[many]{tcolorbox}
\usepackage{xcolor}
\usepackage{listings}
\usepackage{wrapfig}
\newcommand{\ie}{\textit{i}.\textit{e}.}

\begin{document}

\title{TailorMind: Towards Preference-Aligned Multimodal Content Generation}

\author{
  Hengji Zhou$^{1*}$, Ye Liu$^{1*}$, Yufeng Liu$^1$, Si Wu$^1$, 
  Lianghao Xia$^{2\dagger}$, Liqiang Nie$^2$ \\
  $^1$South China University of Technology \\
  $^2$Harbin Institute of Technology, Shenzhen \\
  \texttt{hengjizhou01@gmail.com}, \texttt{202330451251@mail.scut.edu.cn}, \\
  \texttt{202330361751@mail.scut.edu.cn}, \texttt{cswusi@scut.edu.cn}, \\
  \texttt{aka\_xia@foxmail.com}, \texttt{nieliqiang@gmail.com}
}

\renewcommand{\shortauthors}{Zhou et al.}
\def\model{TailorMind}
\def\dataset{TailorBench}

\maketitle
\footnotetext[1]{$^*$Hengji Zhou and Ye Liu have equal contribution to this work.}
\footnotetext[2]{$^\dagger$Lianghao Xia is the corresponding author.}

\begin{abstract}
Personalized content systems depend on available UGC and struggle when suitable content is absent, delayed, or costly to create. Although multimodal generators can synthesize content on demand, how to translate behavioral traces into generation-ready preferences remains underexplored. We study personalized multimodal content generation: creating user-tailored multimodal content without existing item pools or waiting for matching UGC. We propose \textbf{\model}, linking collaborative preference modeling with controllable multimodal generation. \model\ enriches sparse user histories via hypergraph collaborative filtering and optimizes textual profiles with ranking-error feedback and textual gradient descent. Retrieval-augmented style control grounds outputs in authentic UGC patterns, while cross-modal cohesion reflection reduces semantic drift. We construct \textbf{\dataset}, a benchmark from three mainstream platforms evaluated along five dimensions: coherence, novelty,  aesthetic, hallucination, profiling. Experiments show that \model\ achieves competitive or stronger coherence, improves novelty and aesthetic quality over representative generation baselines and ground-truth UGC, demonstrating advantages over retrieving available content or comparable UGC, while achieving up to 29\% Recall gains in reranking. Our code is released at: \url{https://github.com/iLearn-Lab/TailorMind}.
\end{abstract}

\input{intro}
\input{preliminary}

\input{method}
\input{eval}

\input{relate}
\input{conclusion}

\input{limitations}

\input{ethical}

\bibliography{ref}

\appendix
\clearpage
\input{appendix}
\end{document}

%% file: intro.tex
\section{Introduction}
\label{sec:intro}

User-generated content (UGC) has become the dominant source of online media and remains central to content platforms ~\citep{nan2025responsight, safonov2025ntire, li2025ntire}. Yet relying solely on UGC leaves practical gaps. First, novel or technically demanding multimodal content requires specialized skills, tools, time, and production costs, making high-quality large-scale creation difficult for individual creators. Second, emerging topics depend on whether creators notice trends, collect materials, and publish in time, creating a latency gap between user interest and available content. User-aligned artificial intelligence generated content (AIGC) can complement UGC by synthesizing multimodal content on demand, helping platforms cover fresh topics, reduce creation costs, and seed early engagement in specific scenarios.

\begin{figure}
    \centering
    \includegraphics[width=\columnwidth]{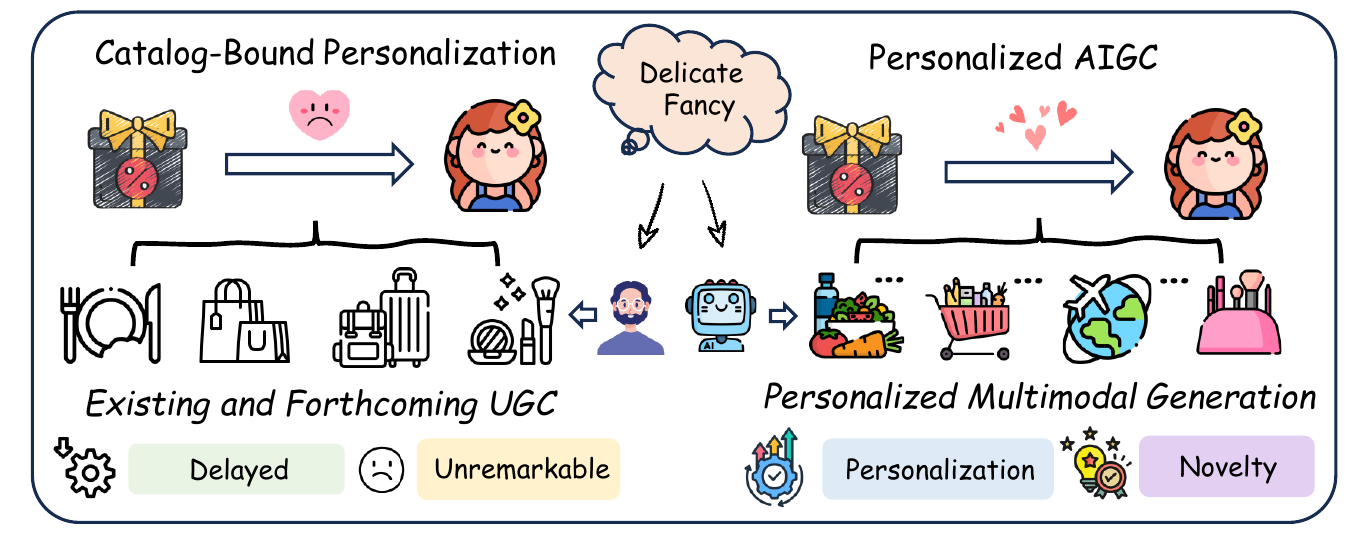}
    \vspace{-0.2in}
    \caption{Personalized multimodal content generation.}
    \label{fig:intro}
    \vspace{-0.15in}
\end{figure}

Recent work increasingly uses large language models (LLMs) as user simulators for synthetic interactions~\citep{Qilin} or ranking agents for preference modeling over item descriptions~\citep{sun2023chatgpt}. Multimodal generative models also enhance item representations via cross-modal synthesis~\citep{pan2025hgdrec, zhang2025collm,fu2024iisan} or augmented views for contrastive learning~\citep{ren2024representation}. These studies show the value of generative models for user understanding, but rarely treat generation itself as the primary personalized output. Closer to our goal, PMG~\citep{shen2024pmg} extracts user preferences from interaction histories as explicit keywords and implicit soft embeddings, jointly conditioning a diffusion model for image generation. Pigeon~\citep{xu2025personalized} instead operates in the visual token space, filtering noisy history images via token-level masking and incorporating multimodal instructions.

However, existing personalization mechanisms, whether catalog-bound or early generation-oriented, retain two \textbf{inherent limitations}. \textbf{Personalization.} Catalog-based methods ~\citep{zhang2024exploring, anand2025survey} only select from available items, so outputs typically \textit{intersect} with user tastes rather than being \textit{fully customized} to unique preference profiles. Recent personalized generation methods condition generators on behavioral descriptions or history-derived visual preferences~\citep{shen2024pmg, xu2025personalized}; yet without optimizing interpretable profiles against real interactions or exploiting collaborative signals to enrich sparse evidence, they struggle with fine-grained, cross-modal, and platform-scale preferences. \textbf{Novelty and immediacy.} Catalog-based methods face a hard novelty ceiling, as recommendations are confined to existing items. When user preferences point to uncovered content—whether driven by long-tail interests~\citep{tang2025one, zhang2025llm} or unexplored preference combinations—systems can only fall back on the nearest catalog approximation. A temporal gap compounds this constraint: platforms must wait for creators to notice demand, produce, and publish, leaving emerging interests chronically underserved.

Overcoming these limitations requires generating content on demand, beyond fixed catalogs and without waiting for future UGC supply, while grounding generation in behavior-derived preferences. This raises two challenges: deriving reliable, actionable profiles from sparse interactions and preserving them throughout multimodal generation.\\\vspace{-0.12in}

\noindent\textbf{C1. Preference-Profile Alignment.} 
Textual profiles should capture nuanced preferences rather than generic summaries. However, sparse and noisy histories make LLM profiling prone to superficial keyword aggregation, while collaborative signals in interaction patterns are easily missed. A reliable solution should optimize profiles against user behavior, requiring feedback and update mechanisms for discrete natural-language representations.
\\\vspace{-0.12in}

\noindent\textbf{C2. Aligning Generated Content with Profiles.} 
Even with an accurate profile, multimodal generation can drift during concept planning, text writing, image synthesis, or video creation. Generated components may partially follow the profile while losing authentic UGC style or cross-modal consistency. Thus, profile constraints must be propagated and checked throughout generation to maintain stylistic authenticity and cross-modal coherence.

To address these challenges, we present \model, a framework for personalized multimodal content generation from user preferences. For C1, \model\ augments sparse histories with hypergraph collaborative filtering and refines textual profiles through textual gradient descent driven by ranking errors on real interactions. For C2, retrieval-augmented style control grounds generation in user-aligned exemplars, while cross-modal cohesion monitoring corrects semantic drift between text and visual modalities. These components generate content that reflects individual preferences while preserving stylistic authenticity. 

Validating personalized generation requires benchmarks with rich multimodal content, real user interactions, and unified evaluation across heterogeneous outputs. We introduce \dataset, a benchmark of real-world interactions and complete multimodal content from mainstream platforms. \dataset\ assesses cross-modal coherence, novelty, aesthetic, content safety, and profiling, enabling evaluation of user alignment and generative fidelity.

Our contributions are summarized as follows:
\begin{itemize}[leftmargin=*, itemsep=2pt, topsep=4pt]
    \item We introduce personalized multimodal generation for synthesizing tailored content beyond the personalization ceiling, novelty boundaries.\vspace{-0.02in}

    \item We propose \model, a personalized generation framework connecting augmented collaborative preference modeling with multimodal generation through data-driven profile optimization and cross-modal alignment mechanisms.\vspace{-0.02in}
    
    \item We construct \dataset, a personalized generation benchmark with rich multimodal content from mainstream platforms and a unified five-dimensional evaluation framework.\vspace{-0.02in}
    
    \item Comprehensive evaluations demonstrate that \model\ achieves superior user alignment, novelty, and multimodal generation quality. 
\end{itemize}

%% file: preliminary.tex
\begin{figure*}[t]
  \centering
  \includegraphics[width=1\linewidth]{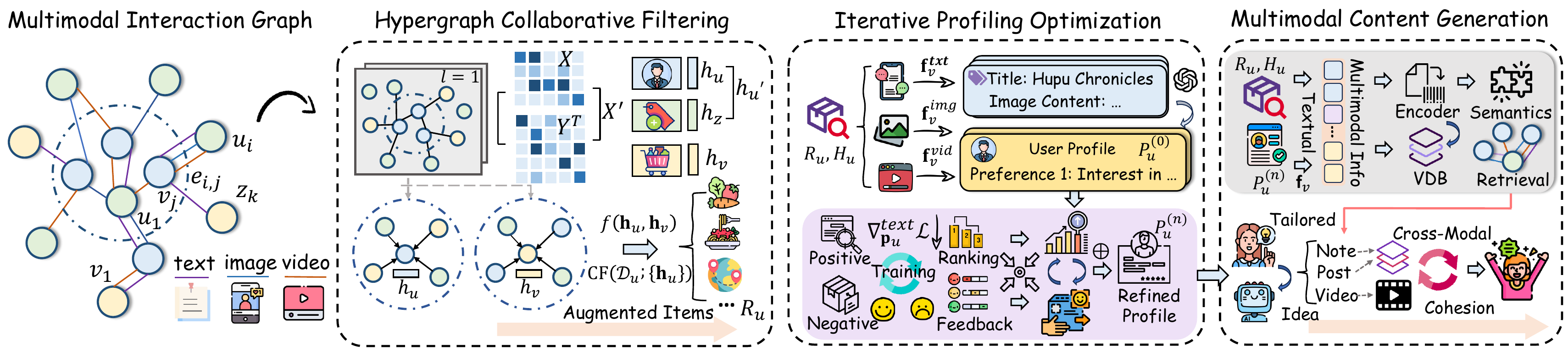}
  \vspace{-0.18in}
  \caption{Overall architecture of the proposed \model\ framework for personalized multimodal generation.} 
  \label{fig:framework}
  \vspace{-0.05in}
\end{figure*}
 \vspace{-0.05in}
\section{Personalized Multimodal Generation}
Personalized multimodal generation bridges two complementary domains: \textbf{user preference modeling}, which infers individual interests from behavioral evidence, and \textbf{multimodal content generation}, which synthesizes multimodal outputs from creative instructions and references.

\noindent\textbf{Task Definition}. Let $\mathcal{U} = \{u_1, \ldots, u_I\}$ be the user set and $\mathcal{V} = \{v_1, \ldots, v_J\}$ the item set. For each user $u \in \mathcal{U}$, the observed interaction history is $\mathcal{D}_u = \{v_{i_1}, \ldots, v_{i_n}\}$. We learn a behavioral preference representation:
\begin{equation}
f_R: (u, \mathcal{D}_u, \mathcal{V}) \rightarrow \mathbf{h}_u,
\end{equation}
where $\mathbf{h}_u$ encodes latent user preferences from historical interactions. Rather than serving as the final output, $\mathbf{h}_u$ provides behavioral evidence for constructing explicit user profiles and creation instructions to synthesize content beyond $\mathcal{V}$. Let $\mathcal{T}$ denote the space of textual creation instructions and $\mathcal{M} = \{\text{image}, \text{text}, \text{video}\}$ the modality set. The generation process is defined as:
{\setlength{\abovedisplayskip}{8pt}
\setlength{\belowdisplayskip}{8pt}
\begin{equation}
f_G: (\mathbf{t}, \{\mathbf{m}_1, \mathbf{m}_2, \ldots\}) \rightarrow \mathcal{O},
\end{equation}
where $\mathbf{t} \in \mathcal{T}$ is the creation instruction, $\{\mathbf{m}_i\}$ are optional multimodal reference materials, and $\mathcal{O} \notin \mathcal{V}$ is the newly synthesized output.\\\vspace{-0.12in}}

\noindent\textbf{Benchmark}. To support personalized generation research, we construct \dataset\ from real-world interactions on \textbf{Rednote}, \textbf{Bilibili}, and \textbf{Hupu}:
{\setlength{\abovedisplayskip}{8pt}
\setlength{\belowdisplayskip}{8pt}
\begin{align}
    \mathcal{B}=\{\mathcal{U},\mathcal{V},\mathcal{D},\mathcal{M}, \mathcal{E}\},
\end{align}}
where $\mathcal{E}$ is a set of evaluation methods assessing five dimensions: 
\textbf{Coherence} (cross-modal consistency),  \textbf{Novelty} (originality relative to existing UGC), \textbf{Aesthetic} (textual and visual quality), \textbf{Hallucination} (absence of fabricated information), and \textbf{Profiling} (quality of user profiles). All user data in \dataset\ is anonymized. Detailed implementation is discussed in Sec.~\ref{sec:exp_setting}.

%% file: method.tex
\section{The Proposed \model\ Framework}
\label{sec:method}
This section elaborates the technical details of the proposed \model\ framework. The overall architecture is depicted in Figure~\ref{fig:framework}.\vspace{-0.05in}

\subsection{Hypergraph Collaborative Filtering}
To generate personalized creation instructions $\textbf{t}$ for user $u$, VLMs require preference-aligned evidence from behavioral history and multimodal content. Directly profiling from $\mathcal{D}_u$ is often noisy and sparse, so \model\ augments user evidence via hypergraph collaborative filtering. We treat content tags as hyperedges because they encode semantic item relations and induce multi-way connections, yielding denser connectivity than user-item bipartite graphs and reducing sparsity degradation.

Formally, let $\mathbf{Y} \in \{0,1\}^{|\mathcal{V}| \times |\mathcal{Z}|}$ denote the item-tag matrix, where $\mathcal{Z}$ is the tag vocabulary, and $\mathbf{X} \in \{0,1\}^{|\mathcal{U}| \times |\mathcal{V}|}$ denote the user-item interaction matrix. We stack $\mathbf{X}$ with $\mathbf{Y}^{\top}$ to construct the augmented adjacency matrix as:
\begin{equation}
    \mathbf{X}' = 
    \begin{bmatrix}
    \mathbf{X} \\[6pt]
    \mathbf{Y}^\top 
    \end{bmatrix}
    \in \{0,1\}^{(|\mathcal{U}|+|\mathcal{Z}|) \times |\mathcal{V}|},
\end{equation}
which defines a unified hypergraph between the joint node set $\mathcal{U} \cup \mathcal{Z}$ and items in $\mathcal{V}$. We initialize embeddings $\mathbf{h}_{u'}^{(0)}$ for each node $u' \in \mathcal{U} \cup \mathcal{Z}$ and $\mathbf{h}_v^{(0)}$ for each item $v \in \mathcal{V}$, then propagate collaborative information through the hypergraph:
\begin{equation}
\mathbf{h}_{u'}^{l+1} = \text{GNN}\left(\mathbf{h}_{u'}^{l}, \{\mathbf{h}_v^{l}: v \in \mathcal{N}(u')\}; \Theta^{l}\right),
\end{equation}
where $\mathcal{N}(u')$ denotes the neighborhood of node $u'$ and $\Theta^{l}$ are trainable parameters at layer $l$. After $L$ propagation layers, we rank unseen items by predicted affinity scores:
{\setlength{\abovedisplayskip}{8pt}
\setlength{\belowdisplayskip}{8pt}
\begin{align}
\hat{y}_{uv} &= f\left(\mathbf{h}_u^{L}, \mathbf{h}_v^{L}\right), \\
\text{CF}(\mathcal{D}_u; \{\mathbf{h}_u\}) &= \mathrm{TopK}\big(\{\hat{y}_{uv}\}_{v \in \mathcal{V} \setminus \mathcal{D}_u}\big).
\end{align}}
where $f$ computes user-item preference scores. Top-$k$ items augment $\mathcal{D}_u$ with collaborative signals, providing broader preference context for profiling.

\subsection{User Profiling with Iterative Optimization}

\subsubsection{Multimodal Profile Initialization}
Existing profiling methods often process modalities independently or concatenate them directly, missing cross-modal semantics that reveal explicit and implicit interests. We instead distill multimodal content into textual descriptions for coherent profile generation. Given item $v$ with multimodal inputs $\{\mathbf{c}_v^m\}_{m \in \mathcal{M}}$, where $\mathcal{M}=\{\text{vid}, \text{img}, \text{txt}\}$, we extract modality-specific textual features:
\begin{equation}
    \mathbf{f}_v^m = \phi_m(\mathbf{c}_v^m), \quad m \in \mathcal{M},
\end{equation}
where $\phi_m(\cdot)$ denotes modality-specific feature extractors based on multimodal models that convert visual and textual content into natural language descriptions. Textual features from the augmented evidence set $\mathcal{D}^{'}_u=\mathcal{D}_u \cup \text{CF}(\mathcal{D}_u)$ are fed into the LLM to synthesize the initial profile:
{\setlength{\abovedisplayskip}{8pt}
\setlength{\belowdisplayskip}{8pt}
\begin{equation}
\mathbf{p}_u^{(0)} = \text{LLM}\left(\{\mathbf{f}_{v}^m : v \in \mathcal{D}^{'}_u\}; ~~p_\text{prof}\right),
\end{equation}}
where $p_\text{prof}$ guides the LLM to reason over multimodal textual features and distill them into a coherent user preference profile $\mathbf{p}_u^{(0)}$, which supports subsequent personalized content generation.

\subsubsection{Textual-Gradient Profile Refinement}
To move beyond heuristic initialization, we formulate profile refinement as a \textbf{feedback-driven optimization process} that minimizes ranking errors on the user's validation ground-truth interaction $\mathcal{Z}_u$ through textual gradient descent, thereby grounding profile updates in actual interactions.
\\\vspace{-0.12in}

\noindent\textbf{Optimization Objective.} 
We optimize the user profile $\mathbf{p}_u$ to maximize ranking quality on interaction groundtruth $\mathcal{Z}_u$ as follows:
{\setlength{\abovedisplayskip}{8pt}
\setlength{\belowdisplayskip}{8pt}
\begin{align}
\label{eq:profile-objective}
\underset{\mathbf{p}_u \in \mathcal{P}}{\min} \; \mathbb{E}_{v^+ \sim \mathcal{Z}_u} \left[ \ell_{\text{rank}}(\mathbf{p}_u, v^+) \right],
\end{align}}
where $\ell_{\text{rank}}(\mathbf{p}_u, v^+)$ measures the rank-position loss of positive interaction item $v^+$ under profile $\mathbf{p}_u$, and $\mathcal{P}$ denotes the profile space of textual descriptions encoding multimodal user preferences.\\\vspace{-0.12in}

\noindent\textbf{Ranking Loss Generation.} 
At iteration $n$, we construct $\mathcal{C}_u$ by sampling one positive item $v^+ \sim \mathcal{Z}_u$ and $M-1$ negative items, then rank them with the current profile $\mathbf{p}_u^{(n)}$ to obtain top-$k$ items:
{\setlength{\abovedisplayskip}{8pt}
\setlength{\belowdisplayskip}{8pt}
\begin{align}
    \mathcal{C}_u = &\{v^+\} \cup \text{Sample}(\mathcal{V} \setminus \mathcal{Z}_u, M-1), \nonumber\\
    &\mathcal{C}_{u,k}^{(n)} = \text{Rank}_{\text{LLM}}\big(\mathbf{p}_u^{(n)}, \mathcal{C}_u\big),
\end{align}}
where $M=100$ in our implementation and $\mathcal{C}_{u,k}^{(n)}$ denotes the top-$k$ ranked items. We compute $\ell_{\text{rank}}(\mathbf{p}_u^{(n)}, v^+)$ from whether $v^+$ appears in top-$k$ and its rank position, yielding a \textbf{discrete feedback signal} that identifies profile aspects that capture user preferences or require adjustment.\\\vspace{-0.12in}

\noindent\textbf{Textual Gradient Descent.} 
Because profiles are textual, conventional numerical gradients are infeasible. Following agent-based optimization~\citep{hu2024adas}, we use a \textbf{textual gradient}: the LLM estimates an update direction from (i) the current profile $\mathbf{p}_u^{(n)}$ and ranking performance, (ii) multimodal features $\{\mathbf{f}_j\}_{j \in \mathcal{C}_{u,k}^{(n)}}$ of top-ranked candidates, and (iii) multimodal features $\{\mathbf{f}_g\}_{g \in \mathcal{Z}_u}$ of groundtruth items. The profile update is:
\begin{equation}
    \label{eq:textual-update}
    \mathbf{p}_u^{(n+1)} = \mathbf{p}_u^{(n)} \oplus \nabla_{\mathbf{p}_u}^{\text{text}} \mathcal{L}(\mathbf{p}_u^{(n)}),
\end{equation}
where $\oplus$ denotes the LLM-based refinement operation incorporating the textual gradient, computed:
\begin{equation}
    \nabla_{\mathbf{p}_u}^{\text{text}} \mathcal{L} = \text{LLM}\big(\mathbf{p}_u^{(n)}, \{\mathbf{f}_j\}_{j \in \mathcal{C}_{u,k}^{(n)}}, \{\mathbf{f}_g\}_{g \in \mathcal{Z}_u}; p_{\text{grad}}\big),
\end{equation}
where $p_{\text{grad}}$ prompts the LLM to improve profile performance. This approximates $\nabla_{\mathbf{p}_u} \mathbb{E}_{v^+ \sim \mathcal{Z}_u}$ through LLM reasoning over ranking loss, analogous to policy-gradient estimation in  reinforcement learning but performed in natural language space.\\\vspace{-0.12in}

\noindent\textbf{Convergence Criterion.} 
The optimization terminates when either (1) ranking loss reaches its optimum, \ie, the groundtruth item $v^+$ ranks first:
\begin{equation}
    \text{rank}(v^+; \mathbf{p}_u^{(n+1)}, \mathcal{C}_u) = 1, \quad \forall v^+ \in \mathcal{Z}_u,
\end{equation}
or (2) the process reaches the maximum iteration $N_{\max}$, ensuring efficiency across users. Thus, \model\ shifts from \textbf{heuristic-based profiling} relying on prompt design to \textbf{data-driven optimization} grounded in groundtruth items, directly improving interaction ranking quality while preserving natural-language interpretability. Gradient optimization prompts are provided in Appendix \ref{sec: prompt templates}\vspace{-0.05in}

\subsection{Personalized Content Generation}
With the optimized user profile $\textbf{p}_u$, \model\ employs specialized generation agents for content categories such as image-text posts and videos. 

\subsubsection{Retrieval-Augmented Style Control}
To control style, \model\ uses retrieval-augmented generation grounded in actual UGC. Although $\mathbf{p}_u$ captures preference semantics, directly translating it into generation instructions may miss stylistic nuances from user interactions. We retrieve top-$k$ exemplars by computing cosine similarity between the profile embedding and multimodal feature representations $\mathbf{f}_i$ of items from historical interactions $\mathcal{D}_u$ and recommended candidates $\text{CF}(\mathcal{D}_u; \{\mathbf{h}_u\})$, defined as :
\begin{equation}
    \mathcal{E}_u^{k} = \operatorname*{\arg\text{top-k}}_{i \in \mathcal{D}_u \cup \text{CF}(\mathcal{D}_u)} \cos(\mathbf{p}_u, \mathbf{f}_i),
\end{equation}
where $\mathcal{E}_u^{k}$ denotes the retrieved exemplar set. These exemplars serve as few-shot references that help the generation agent emulate authentic patterns and stylistic nuances while remaining aligned with preferences encoded in $\mathbf{p}_u$.

\subsubsection{Cross-Modal Cohesion Reflection}
The generation framework aims to align produced multimodal content with $\mathbf{p}_u$. However, long-chain generation across heterogeneous modalities can introduce \textbf{semantic drift}, where generated components diverge from intended profile semantics as abstract preferences are translated into creative concepts and multimodal realizations.

To bridge abstract profile representations and actionable creative directives, we design $R=10$ prototypical idea templates covering creative motifs on contemporary social media platforms. The user profile $\mathbf{p}_u$ is mapped to a style subset $\mathcal{S} \subseteq \{1, \ldots, R\}$ to produce candidate creative ideas $\mathbf{I}_u$:
\begin{equation}
    \mathbf{I}_u = \Psi(\mathbf{p}_u, \mathcal{S})
\end{equation}
where $\Psi(\cdot)$ uses user profiles and selected styles to produce personalized concept prompts. To mitigate semantic drift and enforce profile-content alignment, we employ a \textbf{cross-modal cohesion reflection mechanism}: ClipScore measures consistency between generated visual content $\hat{\mathbf{x}}$ and paired text $\hat{\mathbf{t}}$, while Relevance measures whether $\hat{\mathbf{t}}$ remains consistent with the user's interaction history $\mathcal{D}$ via cosine similarity in text embedding space:
\begin{align}
\mathcal{C}(\hat{\mathbf{x}}, \hat{\mathbf{t}}) &= \text{ClipScore}(\hat{\mathbf{x}}, \hat{\mathbf{t}}), \\
\mathcal{R}(\hat{\mathbf{t}}, \mathcal{D}) &= \text{Relevance}(\hat{\mathbf{t}}, \mathcal{D}),
\end{align}
These cohesion signals, together with $\mathbf{p}_u$ and the previous generation $\mathbf{G}_u^{(n)}$, guide iterative reflective optimization to reduce semantic drift:
\begin{equation}
    \mathbf{G}_u^{(n+1)} = \Phi\big(\mathbf{p}_u, \mathbf{G}_u^{(n)}, \mathcal{C}(\hat{\mathbf{x}}, \hat{\mathbf{t}}), \mathcal{R}(\hat{\mathbf{t}}, \mathcal{D}) \big)
\end{equation}
where $\Phi(\cdot)$ denotes the VLM-based reflective revision function. By maximizing cross-modal cohesion, this mechanism constrains multimodal generation to remain aligned with user preferences.

%% file: eval.tex
\section{Evaluation}
\begin{table}
    \small
    \centering
    \caption{Statistics of the experimental datasets.}
    \label{tab:dataset}
    \setlength{\tabcolsep}{0.9mm}
    \vspace{-0.12in}
    \begin{tabular}{ccccccc}  
      \hline
      Datasets & \#Users & \#Items & \#Tags & \#Text & \#Images & \#Videos\\
      \hline
      Rednote & 12131 & 94997 & 8531 & 7.87 & 7.34 & 2.44\\
      Hupu & 23002 & 126300 & 278 & 6.76 & 11.33 & 1.83\\
      Bilibili & 13349 & 82704 & 8709 & - & - & 6.44 \\ 
      \hline
    \end{tabular}
    \vspace{-0.15in}
\end{table}

We conduct extensive experiments to validate \model's effectiveness across six research questions: \textbf{RQ1} compares \model\ with representative baselines, \textbf{RQ2} examines human preferences over generated content, \textbf{RQ3} analyzes key modules, \textbf{RQ4} studies hyperparameter sensitivity, \textbf{RQ5} evaluates cost and efficiency, and \textbf{RQ6} presents a qualitative case study of generated content.

\begin{table*}[t]
\centering
\renewcommand{\arraystretch}{1.1}
\setlength{\tabcolsep}{0.5mm}
\caption{Generated Content Comparison on Rednote and Hupu Datasets, averaged over Three Runs.}
\label{tab:rednote&hupu-gen}
\small
\vspace{-0.12in}
\begin{tabular}{c|c|cccc|cccc}
    \hline
    \multicolumn{2}{c|}{Model} & \multicolumn{4}{c|}{\textbf{Rednote}} & \multicolumn{4}{c}{\textbf{Hupu}} \\
    \hline
    \multicolumn{2}{c|}{Metric} & 
    Coherence $\uparrow$ & Novelty $\uparrow$ & Aesthetic $\uparrow$  & Hallucination $\downarrow$ & Coherence $\uparrow$ & Novelty $\uparrow$ & Aesthetic $\uparrow$  & Hallucination $\downarrow$ \\
    \hline
    \multirow{7}{*}{\rotatebox{90}{\textbf{I}mage-\textbf{T}ext}} & GPT-4o & 0.6148 & 0.5436 & 0.7243 & 0.0110 & 0.6864 & 0.4723 & 0.6332 & 0.0911 \\
    & Gemini-2.5-flash & 0.6212 & 0.4706 & 0.4391 & 0.2787 & 0.6846 & 0.3826 & 0.2838 & 0.5305 \\
    & Nanobanana pro & 0.6110 & 0.5931 & 0.7262 & 0.0196 & 0.6996 & 0.5529 & 0.6739 & 0.0549 \\
    \cline{2-10}
    & PMG & 0.6155 & 0.5682 & 0.6732 & 0.0550 & 0.6200 & 0.5912 & 0.6572 & 0.2250 \\
    & PIGEON & 0.6050 & 0.6350 & 0.6637 & 0.0124 & 0.6280 & 0.6415 & 0.6706 & 0.0160 \\
    & RAGAR & 0.6255 & 0.5477 & 0.6420 & 0.0215 & 0.6177 & 0.5905 & 0.6584 & 0.0759 \\
    \cline{2-10}
    & Groundtruth-IT & 0.6175 & 0.3458 & 0.3707 & 0.0072 & 0.6147 & 0.2731 & 0.3409 & 0.0197 \\
    \cline{2-10}
    & \textbf{\model-IT} & \textbf{0.6495} & \textbf{0.6643} & \textbf{0.8286} & \textbf{0.0053} & \textbf{0.7132} & \textbf{0.6648} & \textbf{0.7341} & \textbf{0.0100} \\
    \hline
    \hline
    \multirow{6}{*}{\rotatebox{90}{\textbf{V}ideos}} & Veo3.1 & 0.7249 & 0.6675 & 0.8345 & 0.0000 & 0.7228 & 0.6735 & 0.8450 & 0.0000 \\ 
    & Sora2 & 0.7437 & 0.6650 & 0.8295 & 0.0000 &0.7281 & 0.6655 & 0.8410 & 0.0000 \\ 
    \cline{2-10}
    & CIPHER & \textbf{0.7503} & 0.5940 & 0.8040 & 0.0000 & 0.7391 & 0.6915 & 0.8330 & 0.0500 \\
    & PROSE & 0.7429 & 0.6870 & 0.8345 & 0.0000 & \textbf{0.7404} & 0.6905 & 0.8375 & 0.0000 \\
    \cline{2-10}
    & Groundtruth-V & 0.7049 & 0.6731 & 0.7842 & 0.0000 & 0.6231 & 0.5985 & 0.6768 & 0.0000 \\
    \cline{2-10}
    & \textbf{\model-V} & {0.7271} & \textbf{0.6960} & \textbf{0.8430} & \textbf{0.0000} & {0.7230} & \textbf{0.7240} & \textbf{0.8480} & \textbf{0.0000} \\
    \hline
\end{tabular}
\vspace{-0.13in}
\end{table*}

\subsection{Experimental Settings}
\label{sec:exp_setting}
\subsubsection{Datasets and Evaluation Protocols} We construct \dataset\ from three mainstream platforms: \textbf{Rednote}, featuring lifestyle short videos and graphic-text notes; \textbf{Bilibili}, an entertainment-oriented youth video platform; and \textbf{Hupu}, a sports community mainly comprising textual discussions. Table~\ref{tab:dataset} reports detailed statistics. We use leave-two-out splitting for profiling, with the last interaction for testing, the penultimate one for validation, and the rest for optimization. We randomly sample 1,000 users from each dataset~\citep{shen2024pmg}, and compare \model's generated content with baseline outputs and the ground-truth test item along five dimensions: \textbf{Coherence}, where image--text outputs and video-derived textual descriptions are respectively compared against the equal-weighted caption embedding of the user's history and target item using CLIPScore~\citep{hessel2021clipscore} and cosine similarity, reflecting personalization to some extent.  \textbf{Novelty}, where VLMs score novelty and interestingness on a normalized $[0,1]$ scale~\citep{vargas2011rank}; \textbf{Aesthetic}, where VLMs assess textual and visual aesthetics within $[0,1]$~\citep{kao2017deep}; and \textbf{Hallucination}, where web-search tools detect false or fabricated information, labeling hallucinated instances as 1 and otherwise 0. \textbf{Profiling}, where 1:99 negative sampling~\citep{ding2019reinforced} and Recall@N/NDCG@N evaluate reranking performance of user performance modeling.

\subsubsection{Baseline Methods} \model\ is compared with comprehensive baselines: \textbf{i) multimodal AI generation methods:} GPT-4o~\citep{hurst2024gpt}, Gemini-2.5-Flash~\citep{team2023gemini}, Nanobanana pro~\citep{google-nanobanana-pro}, PMG~\citep{shen2024pmg}, Pigeon~\citep{xu2025personalized}, RAGAR ~\citep{ling2026ragar}, Veo3.1~\citep{google-veo31}, Sora2~\citep{openai-sora2}, CIPHER~\citep{gao2024aligning}, and PROSE~\citep{aroca2025aligning}; \textbf{ii) ID-based representation learning methods:} LightGCN~\citep{he2020lightgcn}, BIGCF~\citep{zhang2024exploring}, and IRLLRec~\citep{wang2025intent}; and \textbf{iii) LLM-based textual ranking approaches:}  MoRE~\citep{MoRE}, Re2LLM~\citep{wang2025re2llm}, LLM4Rerank~\citep{gao2025llm4rerank}, and MLLM-MSR~\citep{ye2025harnessing}; \vspace{-0.05in}

\subsubsection{Implementation Details}
All methods are implemented following their original papers. For ID-based representation learning methods, we perform grid search to identify optimal hyperparameters. For long video understanding, we analyze the initial and final 8-second segments separately, then aggregate the information to generate a holistic content description. All LLM-based ranking methods and TailorMind's profile refinement and cohesion reflection use 3 iterations. We employ ViT-B/32 for image encoding and text-embedding-3-small for text encoding. User preferences are extracted by selecting the top-1 preference for personalized multimodal content generation. Our default models are Claude-Sonnet-4.5 for ranking and profiling, Gemini-2.5-Flash-Image for image generation, and Veo-3.1 for video generation. For automatic evaluation only, we use Gemini-2.5-pro as the VLM judge and GPT-5-All for web-based factual verification. \vspace{-0.05in}

\begin{table}[t]
\centering
\renewcommand{\arraystretch}{1.05}
\setlength{\tabcolsep}{0.8mm}
\caption{Generated Content comparison on Bilibili.}
\label{tab:bilibili-gen}
\small
\vspace{-0.12in}
\begin{tabular}{c|cccc}
\hline
Data & \multicolumn{4}{c}{\textbf{Bilibili}} \\
\hline
Metric & Coherence $\uparrow$& Novelty$\uparrow$ & Aesthetic$\uparrow$ & Halluc. $\downarrow$\\
\hline
Veo3.1 & 0.7634 & 0.6710 & 0.8335 & 0.0000\\ 
Sora2 & 0.7932 & 0.6550 & 0.8345 & 0.0000\\ 
\hline
 CIPHER & 0.7800 & 0.6890 & 0.8250 & 0.0000 \\
 Prose & 0.7913 & 0.6225 & 0.7925 & 0.0000 \\
\hline
Groundtruth & 0.7181 & 0.6805 & 0.7802 & 0.0030 \\
\hline
\textbf{\model} & \textbf{0.7947} & \textbf{0.6975} & \textbf{0.8485} & \textbf{0.0000} \\
\hline
\end{tabular}
\vspace{-0.2in}
\end{table}

\subsection{Overall Performance Comparison (RQ1)}

We compare \model\ with baselines from multiple perspectives; results are summarized in Tables \ref{tab:overall retrieval}, \ref{tab:bilibili-gen}, and \ref{tab:rednote&hupu-gen}. \textbf{Stronger User Profiling.} For the profiling dimension, generated user profiles rerank 1:99 candidate sets, and Recall/NDCG measure whether they recover held-out interactions. Table~\ref{tab:overall retrieval} shows that \model\ consistently improves Recall across all datasets, indicating that textual-gradient profile optimization better captures fine-grained preferences from sparse behavioral evidence. \textbf{Higher Novelty and Quality.} Tables~\ref{tab:bilibili-gen} and~\ref{tab:rednote&hupu-gen} show that \model\ generates more novel content than generation baselines and UGC while achieving the best aesthetic scores, demonstrating its ability to synthesize fresh and visually appealing user-aligned outputs across modalities. \textbf{Better Coherence and Reliability.} \model\ achieves higher coherence than UGC and competitive or superior results against generation baselines. Since the coherence reference includes both user history and the target item, these gains indicate stronger history-aware alignment while validating cross-modal cohesion reflection.

\begin{table*}[t]
\centering
\small
\setlength{\tabcolsep}{0.6mm}
\caption{\emph{Reranking} performance comparison, in terms of \textit{Recall} and \textit{NDCG}. Boldface marks the best reranking result. Standard deviations are computed over five runs.}
\label{tab:overall retrieval}
\vspace{-0.12in}
\begin{tabular}{c|cccc|cccc|cccc}
\hline
Data & \multicolumn{4}{c|}{\textbf{Bilibili}} & \multicolumn{4}{c|}{\textbf{Rednote}} & \multicolumn{4}{c}{\textbf{Hupu}} \\
\hline
Metrics & R@10 & N@10 & R@20 & N@20 & R@10 & N@10 & R@20 & N@20 & R@10 & N@10 & R@20 & N@20\\
\hline
LightGCN & 0.3820 & 0.2554 & 0.4780 & 0.2796 & 0.4460 & 0.3605 & 0.4980 & 0.3736 & 0.1160 & 0.0532 & 0.2190 & 0.0791 \\
BIGCF    & 0.3100 & 0.2100 & 0.4120 & 0.2357 & 0.4190 & 0.3575 & 0.4600 & 0.3678 & 0.1130 & 0.0557 & 0.2210 & 0.0825 \\
IRLLRec & 0.4080 & 0.2376 & 0.5240 & 0.2674 & 0.4780 & 0.3446 & 0.5400 & 0.3605 & 0.3320 & 0.1780 & 0.4800 & 0.2155 \\
\hline
MoRE        & 0.1020 & 0.0490 & 0.2050 & 0.0746 & 0.4390 & 0.3172 & 0.5250 & 0.3436 & 0.3010 & 0.1243 & 0.4740 & 0.1678 \\
Re2LLM      & 0.3972 & 0.2633 & 0.5260 & 0.2957 & 0.5460 & 0.3979 & 0.6400 & 0.4216 & 0.4320 & \textbf{0.2100} & 0.6040 & \textbf{0.2532} \\
LLM4Rerank  & 0.3360 & 0.2265 & 0.4470 & 0.2548 & 0.4590 & 0.3415 & 0.5430 & 0.3628 & 0.3870 & 0.1586 & 0.5710 & 0.1894 \\
MLLM-MSR    & 0.3820 & 0.2333 & 0.5150 & 0.2665 & 0.5530 & 0.3812 & 0.6420 & 0.4052 & 0.4170 & 0.2085 & 0.5960 & 0.2420 \\
\hline
\multirow{2}{*}{\textbf{\model}}     & \textbf{0.5260} & \textbf{0.2643} & \textbf{0.6770} & \textbf{0.3023} & \textbf{0.6490} & \textbf{0.4140} & \textbf{0.6800} & \textbf{0.4310} & \textbf{0.4500} & 0.1919 & \textbf{0.6230} & 0.2311 \\
& $\pm$8.2e-4 & $\pm$1.5e-2 & $\pm$2.0e-2 & $\pm$1.3e-2 & $\pm$1.6e-2 & $\pm$1.2e-2 & $\pm$1.3e-2 & $\pm$1.1e-2 & $\pm$1.1e-2 & $\pm$5.1e-3 & $\pm$1.4e-2 & $\pm$5.4e-3\\
\hline
\end{tabular}
\vspace{-0.1in}
\end{table*}

\vspace{-0.1in}

\begin{figure}[t]
  \centering
  
  \includegraphics[width=\linewidth]{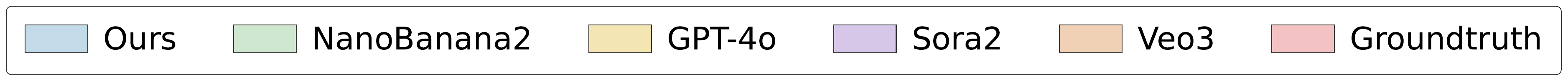}
  \vspace{-0.15in}

  \begin{minipage}[b]{0.31\linewidth}
    \centering
    \includegraphics[width=\linewidth]{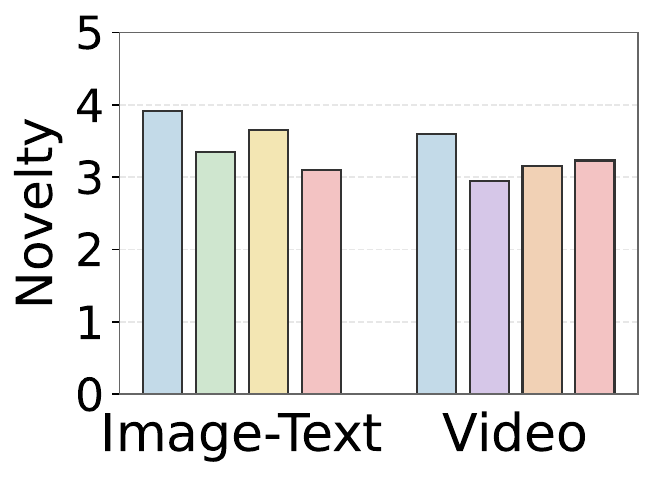}
  \end{minipage}
  \hfill
  \begin{minipage}[b]{0.31\linewidth}
    \centering
    \includegraphics[width=\linewidth]{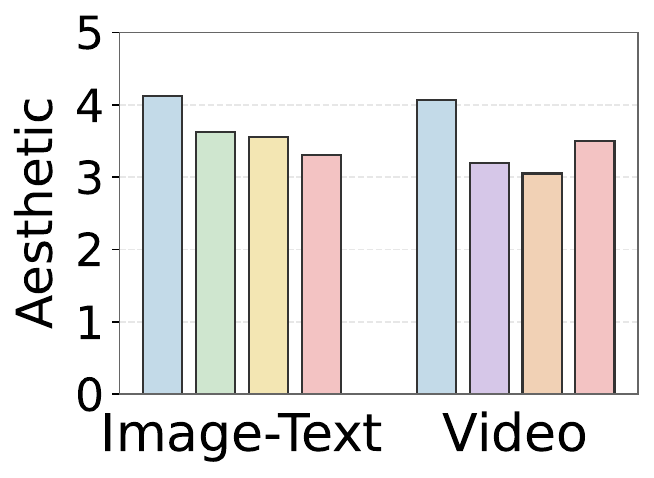}
  \end{minipage}
  \hfill
  \begin{minipage}[b]{0.34\linewidth}
    \centering
    \includegraphics[width=\linewidth]{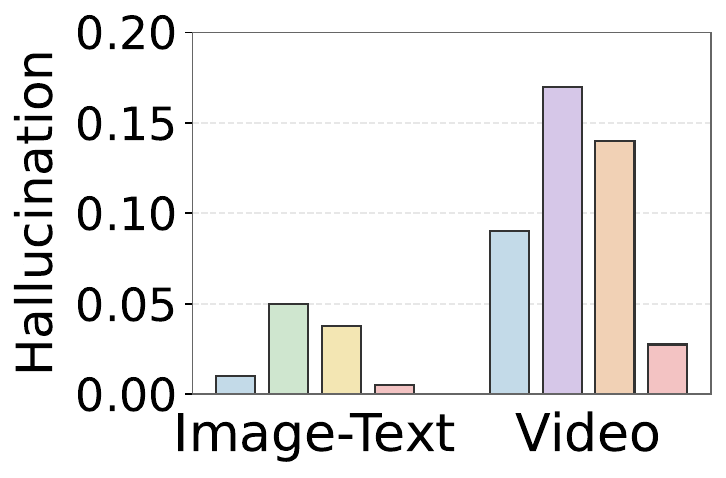}
  \end{minipage}
  \vspace{-0.1in}
  \caption{Human evaluation on created content.}
  \label{fig:human_eval}
  \vspace{-0.15in}
\end{figure}

\subsection{Human Evaluation on Content (RQ2)}
To complement automatic VLM judging and reduce comparison bias with ground-truth UGC, we further conduct a human evaluation of TailorMind’s personalized generation quality. Specifically, we randomly selected 20 generated samples from each image-text and video category and asked 18 volunteers to evaluate anonymized and randomized outputs. Each sample was rated on a 5-point scale for Novelty and Aesthetic quality, while Hallucination was labeled as 1 if misinformation was detected and 0 otherwise; the questionnaire contained 80 items and took about 1.5 hours on average. As shown in Figure~\ref{fig:human_eval}, \model\ achieves stronger novelty and aesthetic quality than generation baselines and ground-truth UGC, while maintaining near-human hallucination performance, validating the effectiveness of our personalized generation.\vspace{-0.07in}

\begin{table}
  \centering
  \small
  \renewcommand{\arraystretch}{1.05}
  \setlength{\tabcolsep}{0.6mm}
  \caption{Performance of different ablated \model.}
  \label{tab:ablation}
  \vspace{-0.12in}
  \begin{tabular}{
      c|c| cc|cc
  }
  \hline
  \multicolumn{2}{c|}{Data} & \multicolumn{2}{c|}{Rednote} & \multicolumn{2}{c}{Hupu}\\
  \hline
  \multicolumn{2}{c|}{Metric} & R@20 & N@20 & R@20 & N@20 \\
  \hline
  \multirow{6}{*}{\rotatebox{90}{\textbf{Profiling}}} & w/o historical & 0.6028 & 0.3207 & 0.5622 & 0.2232 \\
  \cline{2-6}
  & w/o rec & 0.6371 & 0.3543 & 0.5916 & 0.2470\\
  \cline{2-6}
  & w/o TV & 0.6443 & 0.3471 & 0.5842 & 0.2115 \\
  \cline{2-6}
  & w/o IT & 0.2091 & 0.1686 & 0.1777 & 0.0876 \\
  \cline{2-6}
  & w/o IV & 0.6503 & 0.3670 & 0.5965 & 0.2289 \\
  \cline{2-6}
  & \textbf{Origin} & 0.6800 & 0.4310 & 0.6230 & 0.2311 \\
  \hline
  \hline
  \multicolumn{2}{c|}{Metric} & Coherence & Aesthetic & Coherence & Aesthetic \\
  \hline
  \multirow{3}{*}{\rotatebox{90}{\textbf{Gen}}} & w/o RAG & 0.6276 & 0.8020 & 0.6523 & 0.6882 \\
  \cline{2-6}
  &w/o Cohesion & 0.6238 & 0.8253 & 0.6911 & 0.6971 \\  
  \cline{2-6}
  & \textbf{Origin} & 0.6495 & 0.8286 & 0.7132 & 0.7341 \\
  \hline
  \end{tabular}
  \vspace{-0.1in}
\end{table}

\subsection{Ablation Study (RQ3)}
Table~\ref{tab:ablation} quantifies each module's contribution. Removing historical records or hypergraph-recommended items consistently lowers Recall@20 and reduces most NDCG@20 scores, showing that observed behaviors and collaborative signals provide complementary evidence for preference modeling. Ablating TV, IT, or IV inputs further weakens preference profiling, confirming the need for multimodal fusion. For generation, removing RAG reduces coherence and aesthetic quality by weakening UGC-style grounding, while removing Cohesion module hurts cross-modal alignment, indicating its role in limiting semantic drift during long-chain personalized generation.\vspace{-0.06in}

\begin{figure}[t]
  \centering
  \begin{subfigure}[b]{0.236\textwidth}
    \centering
    \includegraphics[width=\linewidth]{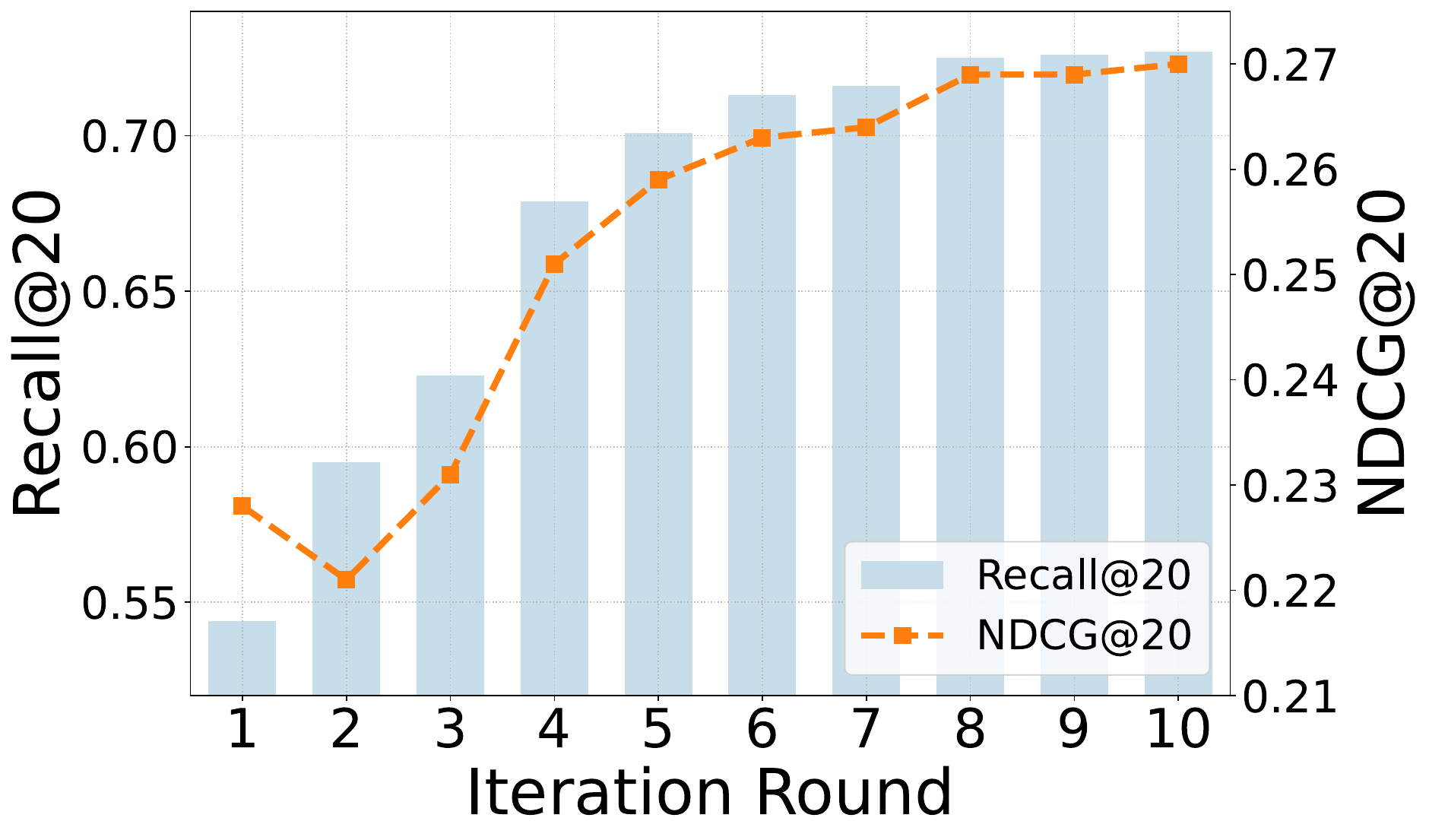}
    \vspace{-0.18in}
    \caption{Profile optimization.}    
    \label{fig:h2}
  \end{subfigure}
  \hfill
  \begin{subfigure}[b]{0.236\textwidth}
    \centering
    \includegraphics[width=\linewidth]{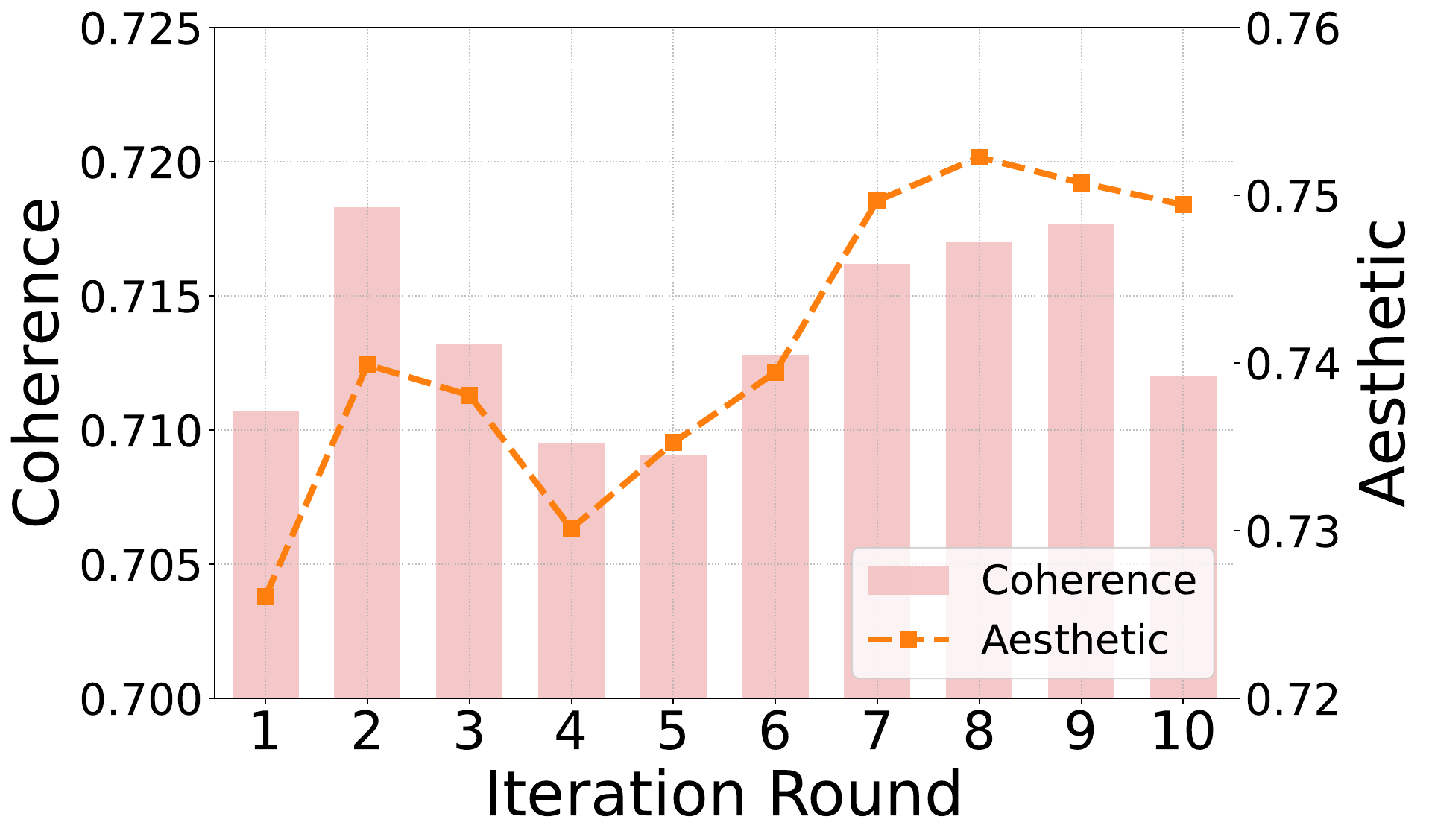}
    \vspace{-0.18in}
    \caption{Generation reflection.}
    \label{fig:h1}
  \end{subfigure}
  \vspace{-0.25in}
  \caption{Hyperparameter study on Hupu dataset.}
  \label{fig:hyper}
  \vspace{-0.2in}
\end{figure}

\subsection{Hyperparameter Study (RQ4)}
Figure~\ref{fig:hyper} reports the effect of profile optimization iterations and generation reflection rounds. For profile optimization, personalization performance rises consistently with more iterations as textual-gradient updates progressively integrate relevant signals and correct mismatches, but gains plateau beyond a certain number of rounds, indicating a practical optimum that balances preference modeling quality and computational cost. For generation reflection, coherence improves notably in early rounds and then stabilizes, while aesthetic quality shows only minor fluctuations throughout, suggesting that a moderate number of reflection rounds is sufficient to reinforce cross-modal consistency.
\vspace{-0.05in}

\begin{figure}
  \centering
 \includegraphics[width=\linewidth]{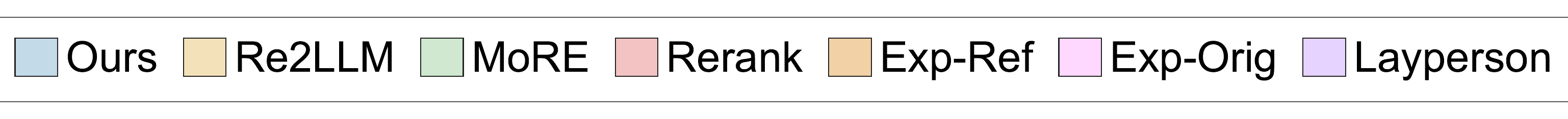}
  \begin{subfigure}[b]{0.29\columnwidth}
    \centering
    \includegraphics[width=\textwidth]{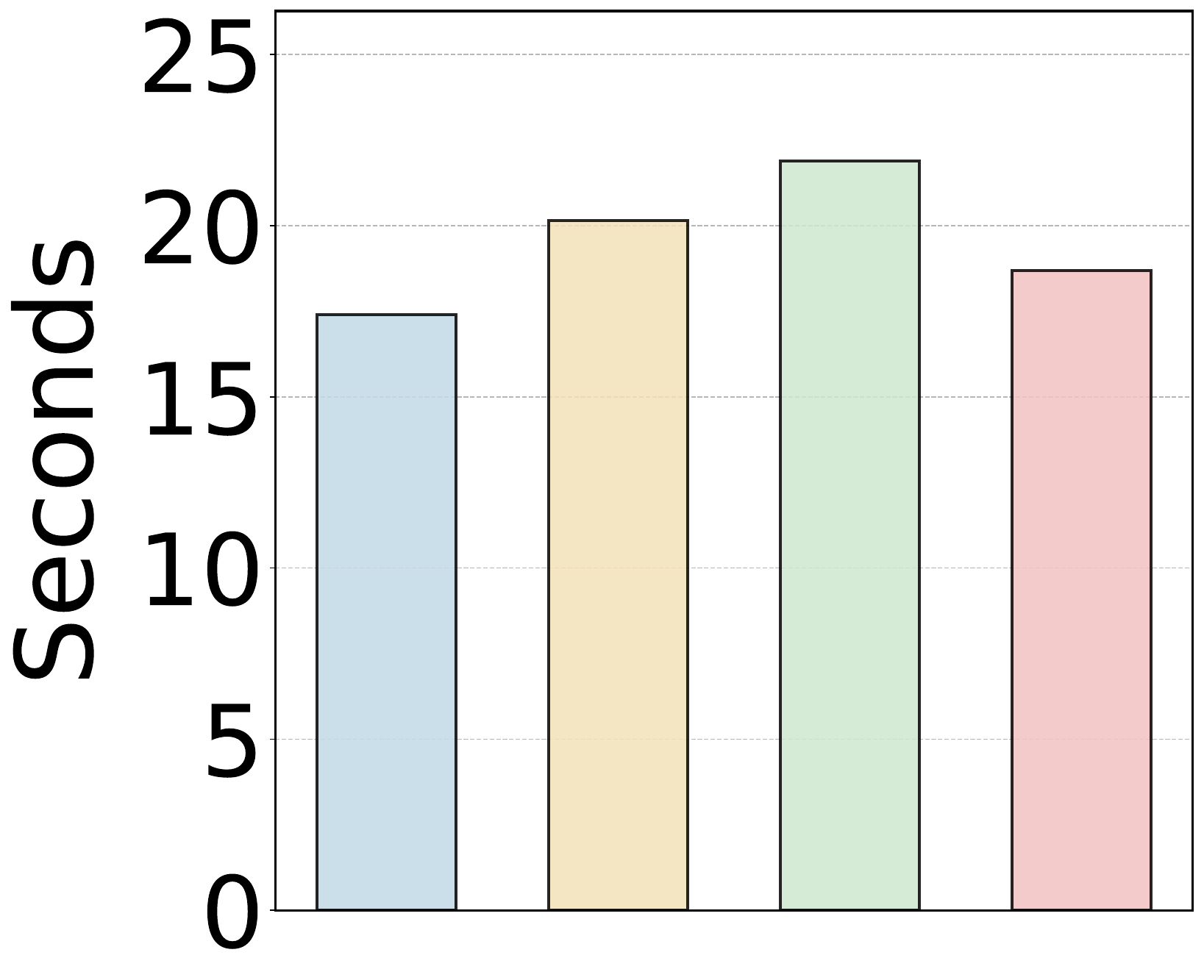}
    \vspace{-0.25in}
    \caption{Profiling time.}
  \end{subfigure}
  \hfill
  \begin{subfigure}[b]{0.32\columnwidth}
    \centering
    \includegraphics[width=\textwidth]{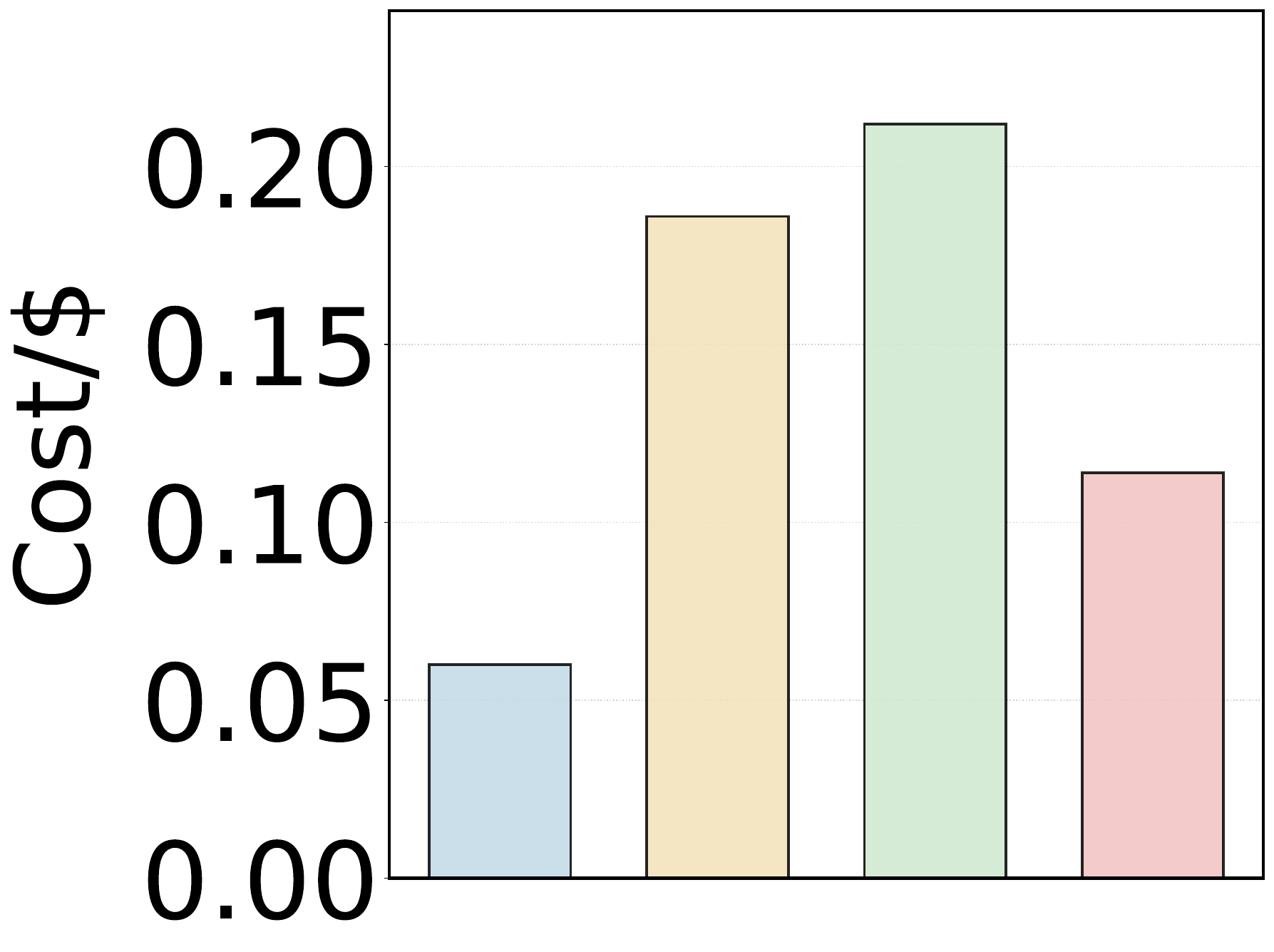}
    \vspace{-0.25in}
    \caption{Profiling costs.}
  \end{subfigure}
  \hfill
  \begin{subfigure}[b]{0.32\columnwidth}
    \centering
    \includegraphics[width=\textwidth]{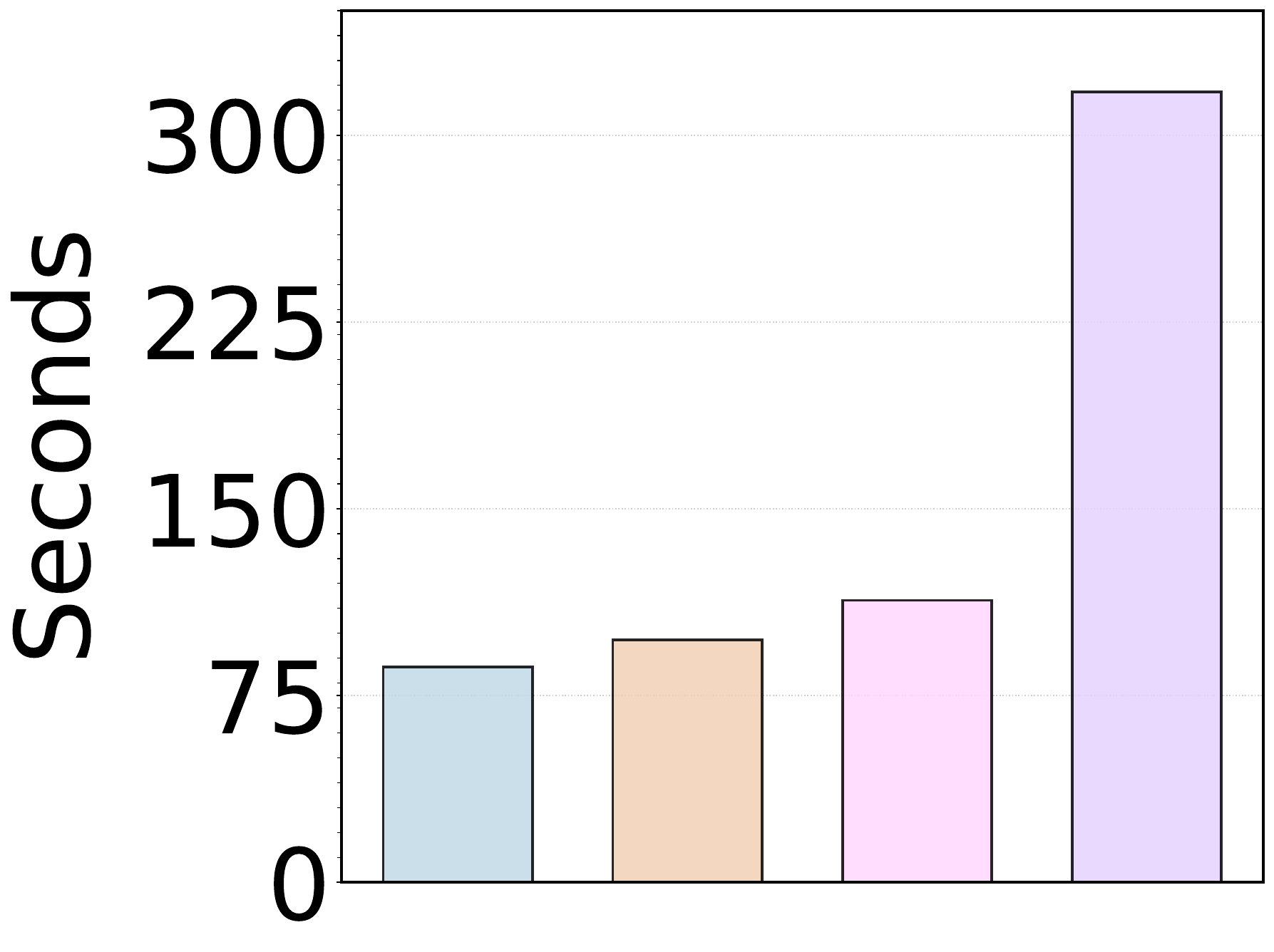}
    \vspace{-0.25in}
    \caption{Generation time.}
  \end{subfigure}
  \vspace{-0.1in}
  \caption{Time and API costs comparison.}
  \label{fig:cost-efficiency}
  \vspace{-0.25in}
\end{figure}

\begin{figure*}[t]
  \centering
  \includegraphics[width=\linewidth]{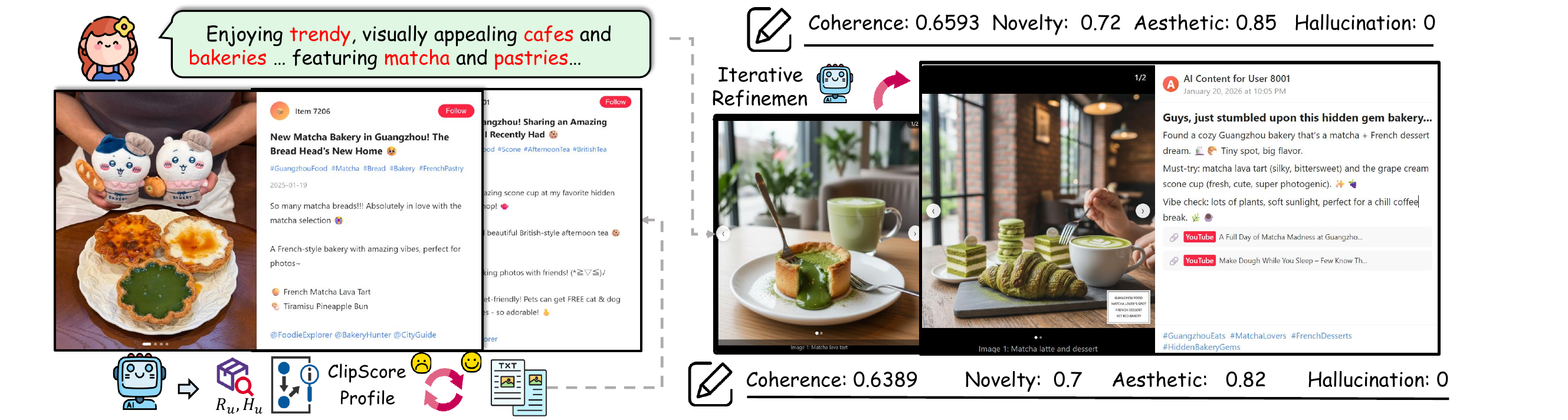}
  \vspace{-0.2in}
  \caption{A RedNote post generated by our \model, personalized for a user who enjoys trendy.}
  \label{fig:casestudy}
  \vspace{-0.2in}
\end{figure*}

\subsection{Cost-Efficiency Study (RQ5)}

We evaluate the API costs and time efficiency of \model\ in Figure~\ref{fig:cost-efficiency}. For the profiling process, \model\ reduces average computational overhead and latency compared to LLM-based baselines while achieving superior performance. For the content generation process, we compare \model\ against three human creator groups sampled from 10 Rednote user profiles: a professional team manually operating the same AI tools as \model\ from scratch (Expert-Orig), the same expert team refining existing groundtruth UGC with those tools (Expert-Ref), and 8 ordinary users without professional creation experience (Layperson). All reported costs and times are averaged across users. \model\ substantially reduces generation time over all three groups, demonstrating strong suitability for practical personalized content generation. 

\subsection{Case Study (RQ6)}
Figure~\ref{fig:casestudy} shows a Rednote user (ID 8001) and several genuine interactions. From these multimodal histories, \model\ identifies a preference for street eats, trendy cafes, and visually appealing food photography. The generated image-text post preserves these concrete interests while producing new content beyond the observed UGC items. Guided by cross-modal cohesion reflection, the post improves Coherence, Novelty, and Aesthetic Quality while maintaining zero Hallucination, indicating that iterative reflection corrects cross-modal mismatch without weakening history-aware alignment. The final content exhibits strong visual-textual alignment, supported by textual-gradient profile refinement, retrieval-augmented UGC style control, and cross-modal cohesion reflection.

%% file: relate.tex
\vspace{-0.1in}
\section{Related Work}
\label{sec:relate}

\noindent\textbf{Personalized Generation}.
Personalized generation adapts content to users, styles, or preference descriptions. PMG~\citep{shen2024pmg} derives personalized multimodal generation instructions with LLMs, while Pigeon~\citep{xu2025personalized} aligns image generation with visual preferences inferred from histories and instructions. Related preference-alignment methods infer user intent from edits or demonstrations to steer generated outputs~\citep{gao2024aligning,aroca2025aligning}. Yet existing methods often rely on descriptive preference signals, with limited correction for semantic drift. \model\ instead optimizes profiles with interaction feedback and maintains profile-content alignment through retrieval-augmented control and cross-modal cohesion reflection.\\\vspace{-0.12in}

\noindent\textbf{Multimodal Generative Models}.
Recent AIGC models seamlessly synthesize diverse, high-quality text, images, and videos from free-form natural-language instructions. Large vision-language foundation models increasingly support multimodal reasoning and cross-modal content creation~\citep{hurst2024gpt,jin2025large}, while image and video generation models steadily improve visual fidelity, prompt following, temporal consistency, and motion control~\citep{sun2025content,seedance2026seedance}. These advances enable interactive user-facing content, but are still typically optimized for prompt-conditioned generation rather than behavior-grounded personalization, making noisy, sparse histories difficult to translate into reliable generation conditions.\\
\vspace{-0.1in}

\noindent\textbf{Recommendation Systems}.
Recommendation systems model user preferences from behavioral data. Traditional recommenders capture collaborative signals through graph or intent-based representations~\citep{he2020lightgcn, zhang2024exploring,zhao2025symmetric}, while sequential and multimodal recommenders model temporal dynamics and heterogeneous content features~\citep{ren2024representation,fu2025efficient,lin2025contrastive}. Recent LLM-based ranking methods use language models to understand descriptions, rerank candidates, simulate users, or refine recommendations~\citep{sun2023chatgpt,MoRE,ye2025harnessing}. These methods mainly select from existing catalogs; our work instead uses recommendation signals to build generation-ready profiles and synthesize new personalized content.

%% file: conclusion.tex
\vspace{-0.15in}
\section{Conclusion}
This paper studies personalized multimodal content generation beyond existing UGC catalogs and proposes \model, which connects behavior-grounded preference modeling with controllable generation through hypergraph collaborative filtering, textual-gradient profile optimization, retrieval-augmented style control, and cross-modal cohesion reflection. We further construct \dataset, a benchmark with user interactions, multimodal content from three mainstream platforms, and five-dimensional evaluation metrics. Experiments show that \model\ improves preference alignment and generates higher-quality personalized content than generation and retrieval baselines. Future work will explore interactive generation, stronger safety control and more efficient profile optimization.

%% file: limitations.tex
\section*{Limitations}
Current social platforms are still largely driven by user-generated content, and creators may not have access to explicit or sufficiently structured follower preferences when producing personalized multimodal content. This limits how directly AIGC systems can support creators in matching audience interests. If platforms can disclose partial, aggregated, high-level follower preference signals to creators within acceptable privacy, consent, and platform-governance constraints, such signals may better facilitate AIGC-assisted creation and reduce the time and effort required for content production.

%% file: ethical.tex
\section*{Ethical Considerations}
The datasets used in this study are collected from publicly available user-generated content on three mainstream platforms. We anonymize all user-related information, remove personally identifiable information, and retain only internal identifiers needed to link interaction histories with items. All data are used solely for aggregate-level, privacy-preserving research evaluation, and we do not attempt to re-identify or target individual users.

%% file: appendix.tex
\section{Appendix}

\subsection{Baseline Methods}
  To ensure a comprehensive study, we compare TailorMind against 17 baseline methods including various techniques for personalized content generation.
 
  \textbf{Multimodal AI Generation Methods}
  \setlength{\leftmargini}{10pt}
  \begin{itemize}
  \item {\textbf{GPT-4o}} \cite{hurst2024gpt}: It adopts an end-to-end omni-architecture for low-latency multimodal understanding and generation.
  \item {\textbf{Gemini-2.5-Flash}} \cite{team2023gemini}: Built on native multimodal pretraining and MoE scaling, it supports efficient cross-modal reasoning.
  \item {\textbf{Nanobanana pro}} \cite{google-nanobanana-pro}: It uses reasoning-enhanced diffusion with controllable visual grounding for coherent high-resolution image generation from 1K to 4K outputs.
  \item {\textbf{PMG}} \cite{shen2024pmg}: LLM-derived user signals are used to guide personalized multimodal content generation effectively.
  \item {\textbf{Pigeon}} \cite{xu2025personalized}: Retrieval-augmented preference inference guides frozen agents without model fine-tuning.
  \item {\textbf{RAGAR}} \cite{ling2026ragar}: Semantic retrieval prioritizes relevant historical records, whereas ranking feedback jointly accounts for personalized adaptation and faithful representation.
  \item {\textbf{Veo3.1}} \cite{google-veo31}: Diffusion-based video generation is combined with audio synthesis and narrative-aware conditioning.
  \item {\textbf{Sora2}} \cite{openai-sora2}: A spatiotemporal diffusion transformer enables synchronized video-audio generation with consistent motion.
  \item {\textbf{CIPHER}} \cite{gao2024aligning}: Historical user edits are retrieved to infer preferences and align generated outputs with user intent.
  \item {\textbf{PROSE}} \cite{aroca2025aligning}: Iterative refinement and consistency checks infer user preferences from demonstrations.
  \end{itemize}

  \textbf{ID-based Representation Learning Methods}
  \setlength{\leftmargini}{10pt}
  \begin{itemize}
  \item {\textbf{LightGCN}} \cite{he2020lightgcn}: Simplified graph convolution propagates user-item embeddings for collaborative filtering.
  \item {\textbf{BIGCF}} \cite{zhang2024exploring}: Behavior-aware graph modeling enhances collaborative filtering with richer user-item signals.
  \item {\textbf{IRLLRec}} \cite{wang2025intent}: A dual-tower design aligns LLM-derived textual intents with interaction-based preferences.
  \end{itemize}

  \textbf{LLM-based Textual Ranking Approaches}
  \setlength{\leftmargini}{10pt}
  \begin{itemize}
  \item {\textbf{MoRE}} \cite{MoRE}: Multiple LLM reflectors capture explicit, implicit, and collaborative preferences, while a meta-reflector selects the most suitable perspective.
  \item {\textbf{Re2LLM}} \cite{wang2025re2llm}: LLM self-reflection generates hints, while PPO-based retrieval selects useful guidance.
  \item {\textbf{LLM4Rerank}} \cite{gao2025llm4rerank}: Aspect-graph reasoning with Chain-of-Thought supports personalized item reranking.
  \item {\textbf{MLLM-MSR}} \cite{ye2025harnessing}: Multimodal item understanding is paired with temporal behavior modeling for recommendation.
  \end{itemize}

\begin{table}
  \centering
  \small
  \renewcommand{\arraystretch}{1.15}
  \setlength{\tabcolsep}{0.35mm}
  \caption{Comparison on different backbones.}
  \label{tab:backbone}
  \vspace{-0.12in}
  \begin{tabular}{
      c|c|c|c|c|c|c
      }
  \hline
  Backbone & \multicolumn{2}{c|}{Gemini-3-flash} & \multicolumn{2}{c|}{DeepSeek-V3.2} & \multicolumn{2}{c}{GPT-5.2} \\
  \hline
    Metric & R@20 & N@20 & R@20 & N@20 & R@20 & N@20  \\
  \hline
  Re2LLM & 0.4060 & 0.2512 & 0.4610 & 0.2655 & 0.5190 & 0.2891  \\
  \hline
  MLLM-MSR & 0.5800 & 0.3524 & 0.5320 & 0.2865 & 0.6050 & 0.3553  \\
  \hline
  \textbf{\model} & 0.7760 & 0.3754 & 0.7780 & 0.3548 & 0.7900 & 0.3608 \\
  \hline
  \hline
  Backbone & \multicolumn{2}{c|}{IRLLRec} & \multicolumn{2}{c|}{RLMRec} & \multicolumn{2}{c}{LightGCN} \\
  \hline
    Metric & R@20 & N@20 & R@20 & N@20 & R@20 & N@20  \\
  \hline
  \textbf{\model} & 0.7210 & 0.2960 & 0.7350 & 0.3021 & 0.6770 & 0.3023\\
  \hline
  \end{tabular}
  \vspace{-0.1in}
\end{table}

\vspace{-0.05in}
\subsection{Impact of Backbones (RQ7)}
To isolate the effect of backbone architecture for ranking, we substitute different backbones (recommendation models and LLMs) and evaluate personalization on Bilibili data using Recall@20 and NDCG@20 (Table~\ref{tab:backbone}).

We find LLMs with stronger contextual understanding and richer semantics produce more informative user profiles, boosting ranking. GPT-5.2 leads, while DeepSeek-V3.2 offers competitive results with better efficiency. Gemini-3-flash is competitive; LightGCN is a solid baseline. Advanced recommenders like IRLLRec and RLMRec, which integrate collaborative information, consistently improve ranking. These results validate \model's robustness and adaptability. The synergy between LLMs and collaborative recommenders consistently boosts performance across all backbone combinations.

\begin{figure*}[p]
    \centering
    \captionsetup{type=figure}
    \begin{minipage}{\linewidth}
        \includegraphics[width=\textwidth]{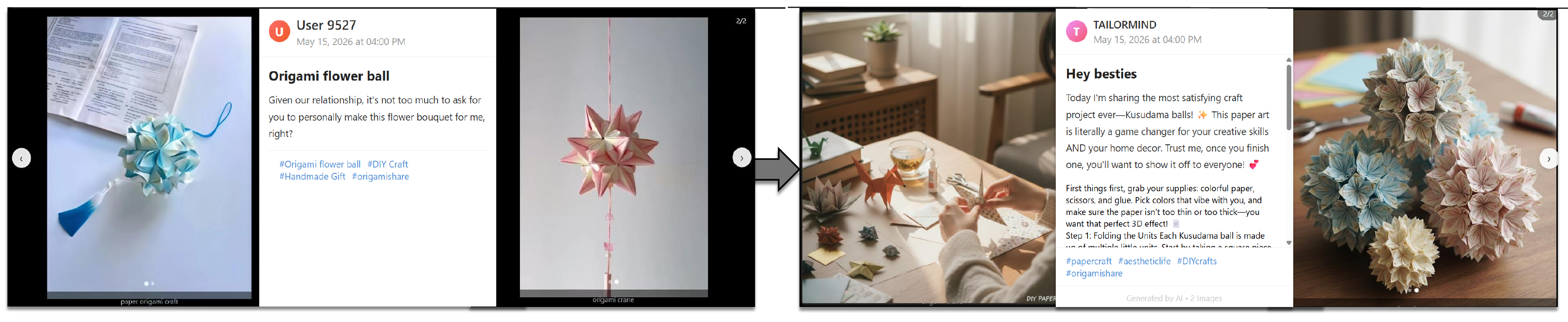}
        \label{fig:appx_casestudy1}
    \end{minipage}
    \begin{minipage}{\linewidth}
        \vspace{-0.13in}
        \includegraphics[width=\textwidth]{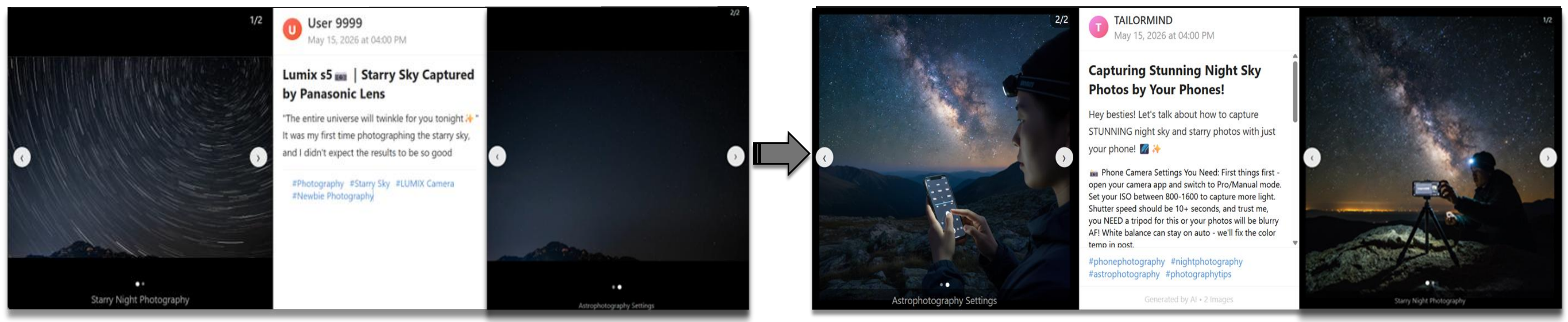}
        \label{fig:appx_casestudy2}
        \vspace{-0.2in}
        \subcaption{Rednote: Image-Text}
    \end{minipage}
    \begin{minipage}{\linewidth}
        \vspace{0.01in}
        \includegraphics[width=\textwidth]{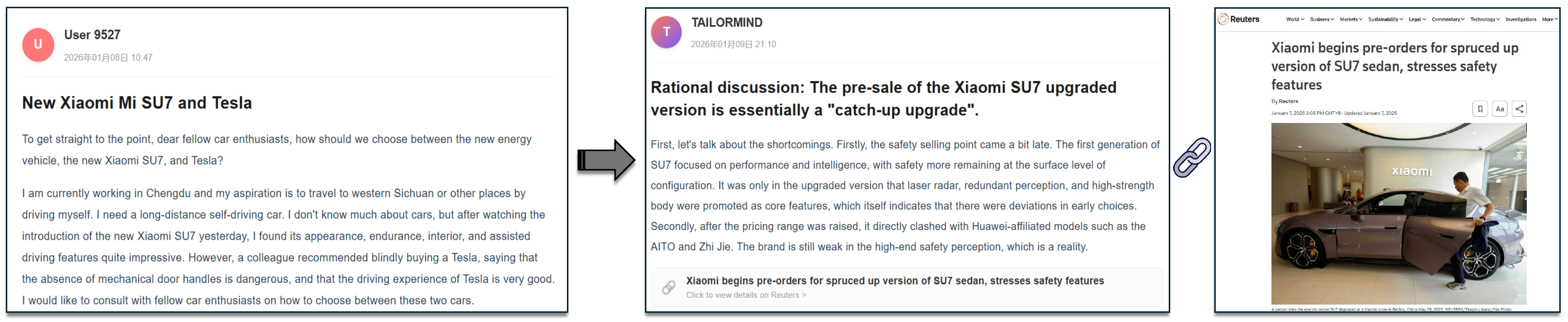}
        \label{fig:appx_casestudy3}
    \end{minipage}
    \begin{minipage}{\linewidth}
        \vspace{-0.08in}
        \includegraphics[width=\textwidth]{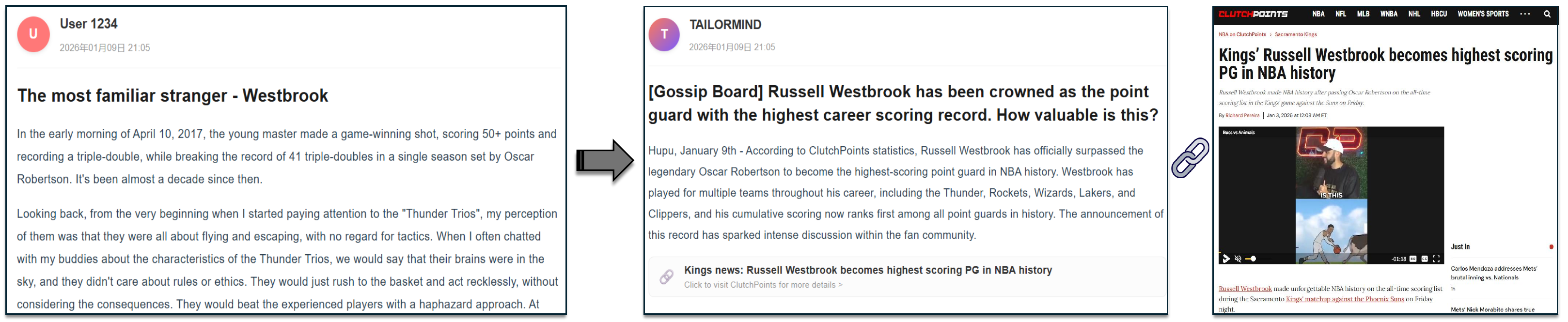}
        \label{fig:appx_casestudy4}
        \vspace{-0.2in}
        \subcaption{Hupu: Link-Post}
    \end{minipage}
    \begin{minipage}{\linewidth}
        \vspace{0.01in}
        \includegraphics[width=\textwidth]{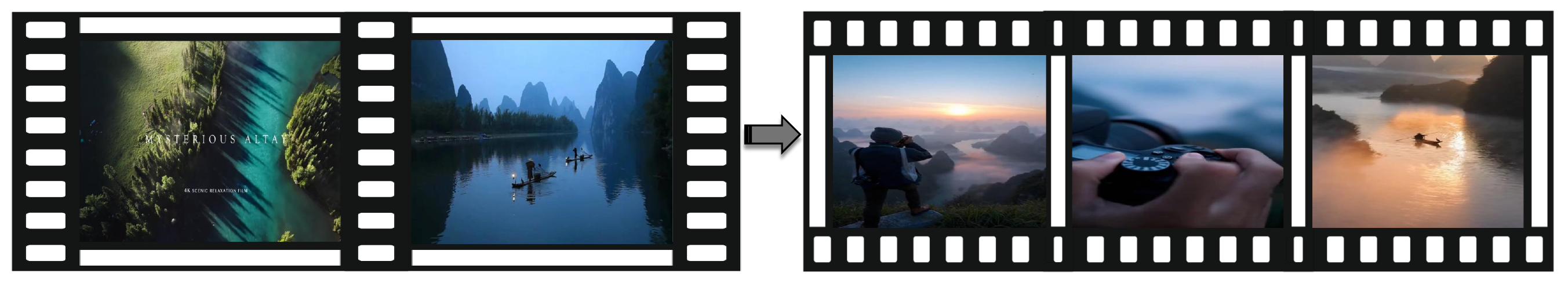}
        \label{fig:appx_casestudy5}
    \end{minipage}
    \begin{minipage}{\linewidth}
        \vspace{-0.18in}
        \includegraphics[width=\textwidth]{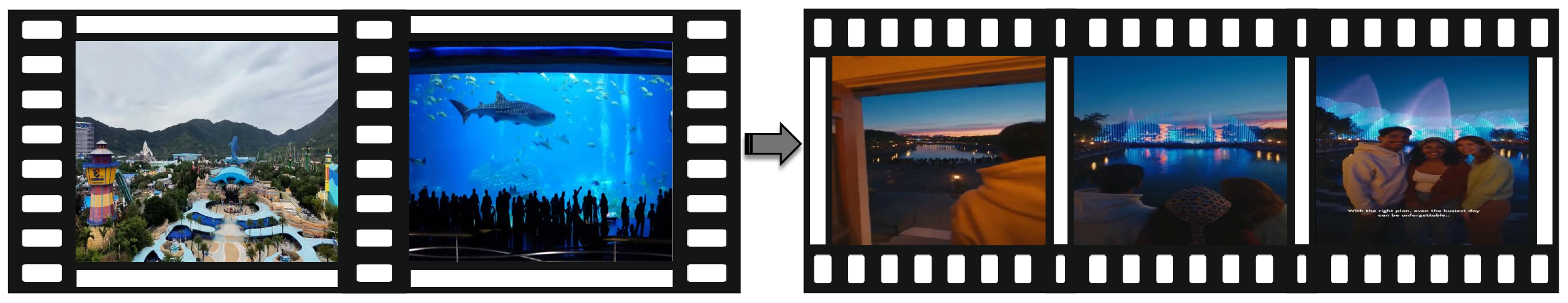}
        \label{fig:appx_casestudy6}
        \vspace{-0.24in}
        \subcaption{Bilibili: Video}
    \end{minipage}
    \vspace{-0.05in}
    \caption{TailorMind cases on three platforms: user history (left) vs. preference-based generated products (right).}
    \label{fig:overall_case_studies}
\end{figure*}
\clearpage

\subsection{Human evaluation}
For human evaluation, 18 student volunteers recruited from online communities rated anonymized image-text and video outputs generated by different methods, with all items randomly shuffled to conceal method identities. For the cost-efficiency study, we involved a professional image-text creation team with regular experience producing image-text content for Rednote, together with 8 general student users. All participants remained anonymous, and all generated content and ratings were used solely for academic research purposes.

\

\subsection{Dataset Details}
\textbf{Dataset Description.} We collected publicly available data from three platforms using a specialized crawler, anonymized all user-related information, and retained unique identifiers, tags, and timestamps for temporal and structural consistency. (i) \textbf{Rednote} metadata includes user/item IDs, content type (image-text/video), text, tags, and media files. (ii) \textbf{Hupu} data captures forum post titles, textual content, multimodal assets, and user replies. (iii) \textbf{Bilibili} entries encompass video-centric features such as duration, introduction, and semantic titles of user-created favorite folders. 

\textbf{Evaluation Sampling.} For each dataset, we randomly sampled 1,000 users as evaluation pool for retrieval and generation studies. For image-text generation, we produced one post for each sampled Rednote and Hupu user but omitted Bilibili because it is predominantly video-based; for video generation, considering cost and latency, we sampled 50 users from dataset's 1,000-user pool. Generated videos have durations ranging from 8 to 45 seconds.
\vspace{-0.20in}

\subsection{Prompt Templates}
\label{sec: prompt templates}
\begin{figure}[t] 
    \vspace{-4mm}
    \centering
    \includegraphics[width=\linewidth]{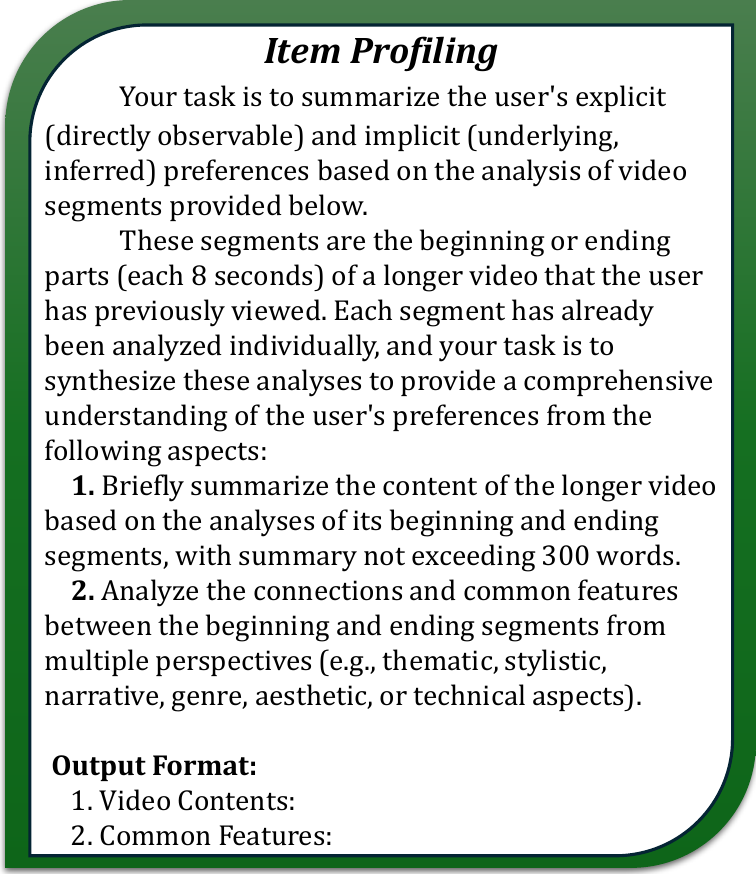} 
    \caption{Video-item variant of the Item Profiling.}
    \label{fig:item_profiling}
\end{figure}

\begin{figure}[h] 
    \vspace{-12mm}
    \centering
    \includegraphics[width=\linewidth]{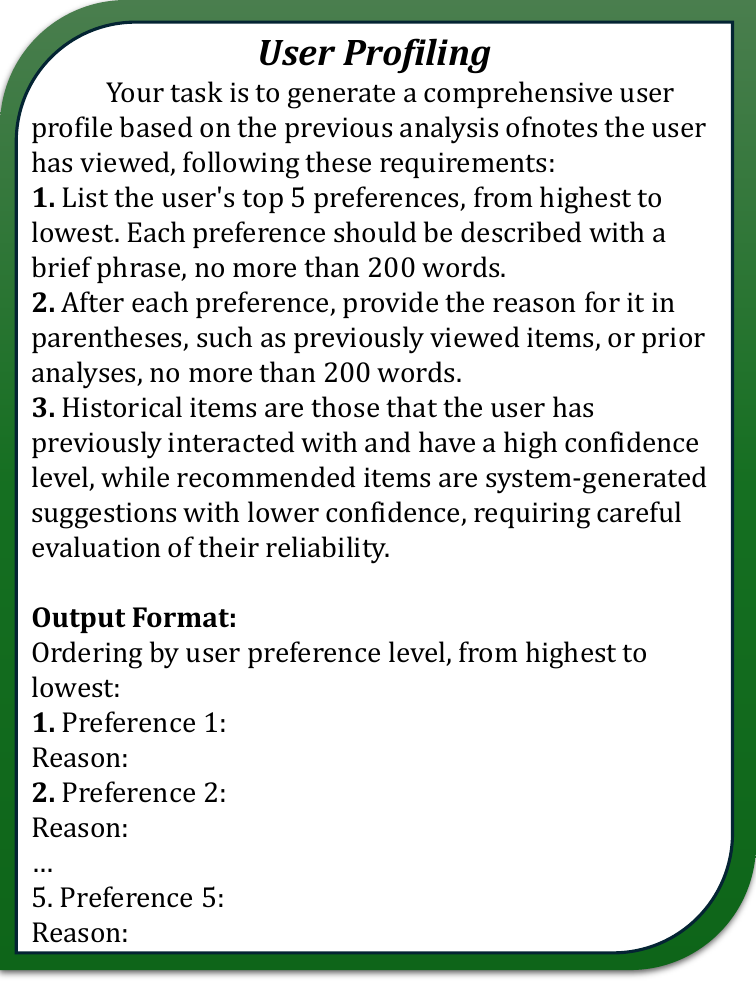} 
    \caption{User Profiling prompt for aggregating item-level profiles into user personas.}
    \label{fig:user_profiling}
\end{figure}

\begin{figure}[t]
    \centering
    \begin{subfigure}{\linewidth}
        \centering
        \includegraphics[width=\linewidth]{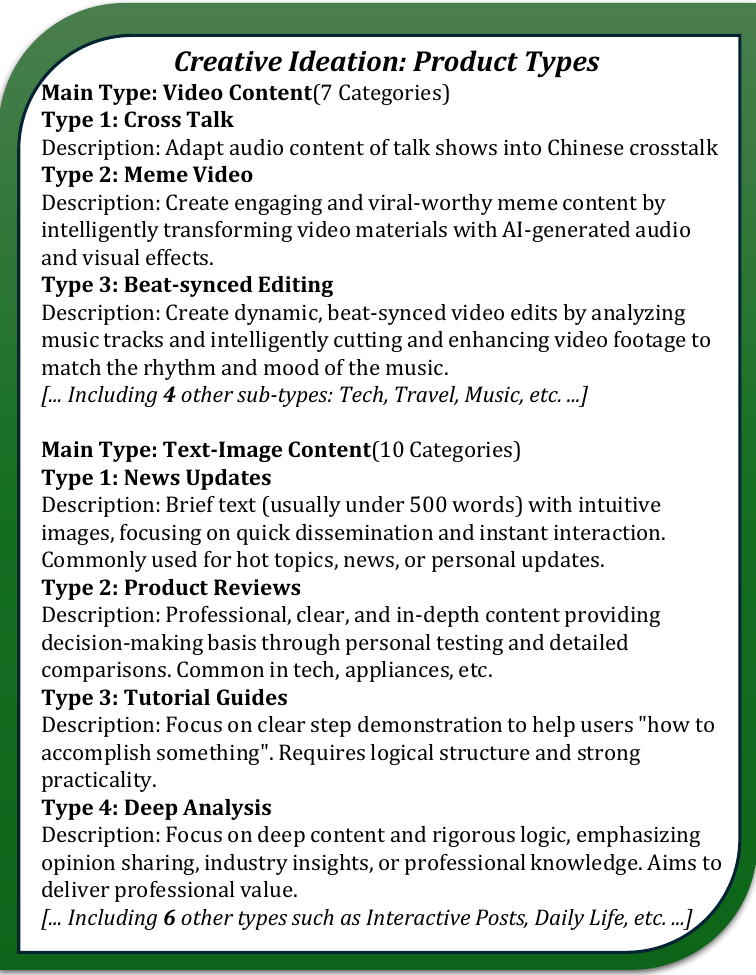}
        \vspace{-2mm}
        \caption{Product Types}
        \label{fig:product_types}
    \end{subfigure}

    \vspace{2mm}

    \begin{subfigure}{\linewidth}
        \centering
        \includegraphics[width=\linewidth]{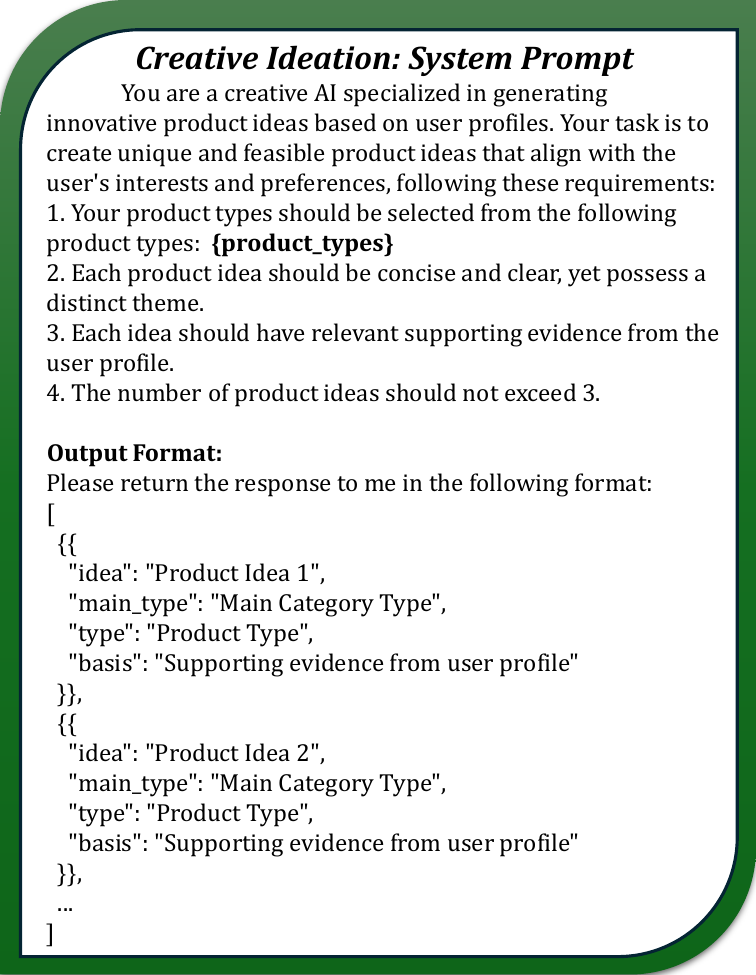}
        \vspace{-2mm}
        \caption{Creative Ideation Prompt}
        \label{fig:ideation}
    \end{subfigure}
    
    \caption{The creative ideation stage: predefined product categories (a) and the core prompt template (b).}
    \label{fig:creative_ideation}
\end{figure}

\begin{figure}[t]
    \centering
    \begin{subfigure}{\linewidth}
        \centering
        \includegraphics[width=\linewidth]{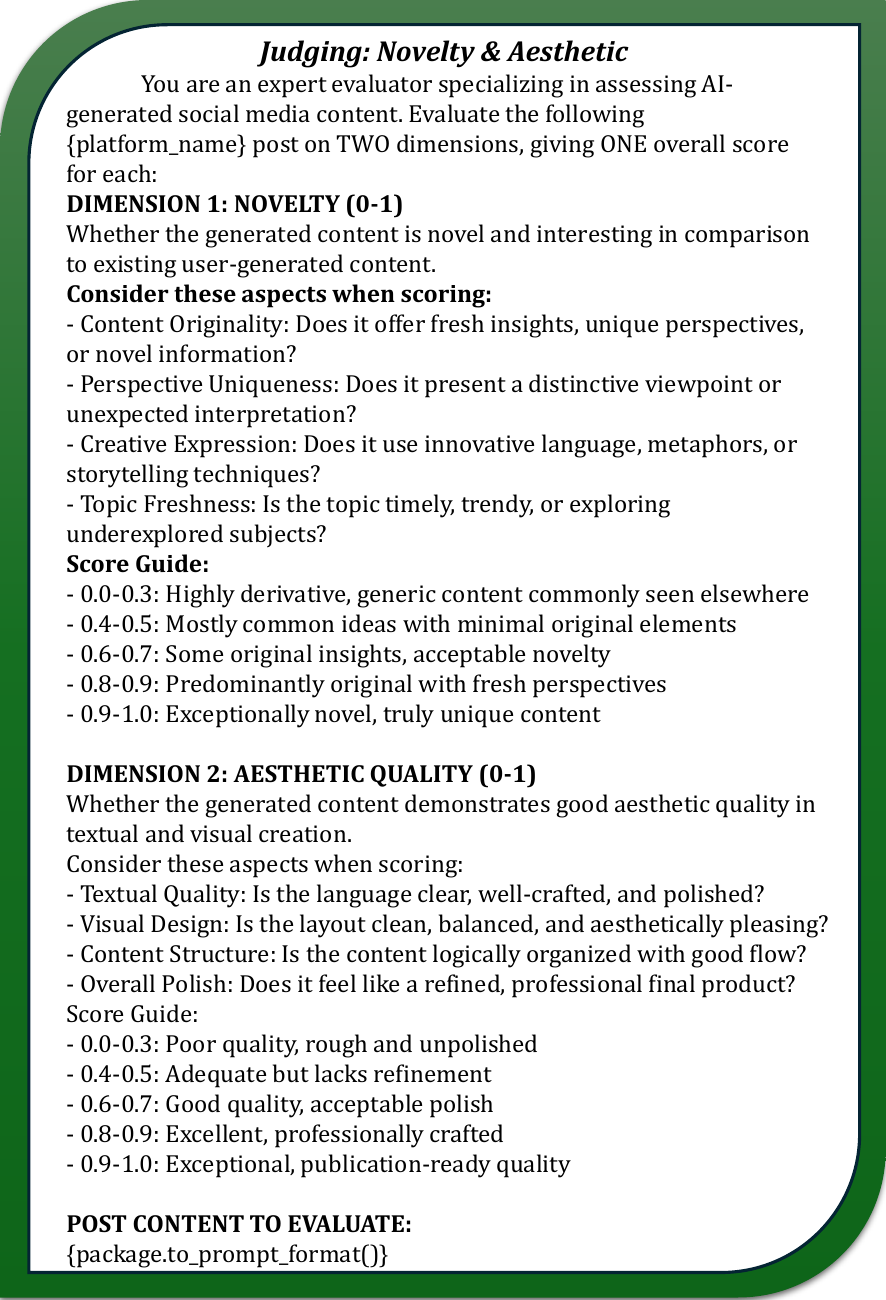}
        \caption{Novelty \& Aesthetic Evaluation}
        \label{fig:judging_novelty}
    \end{subfigure}

    \vspace{2mm}

    \begin{subfigure}{\linewidth}
        \centering
        \includegraphics[width=\linewidth]{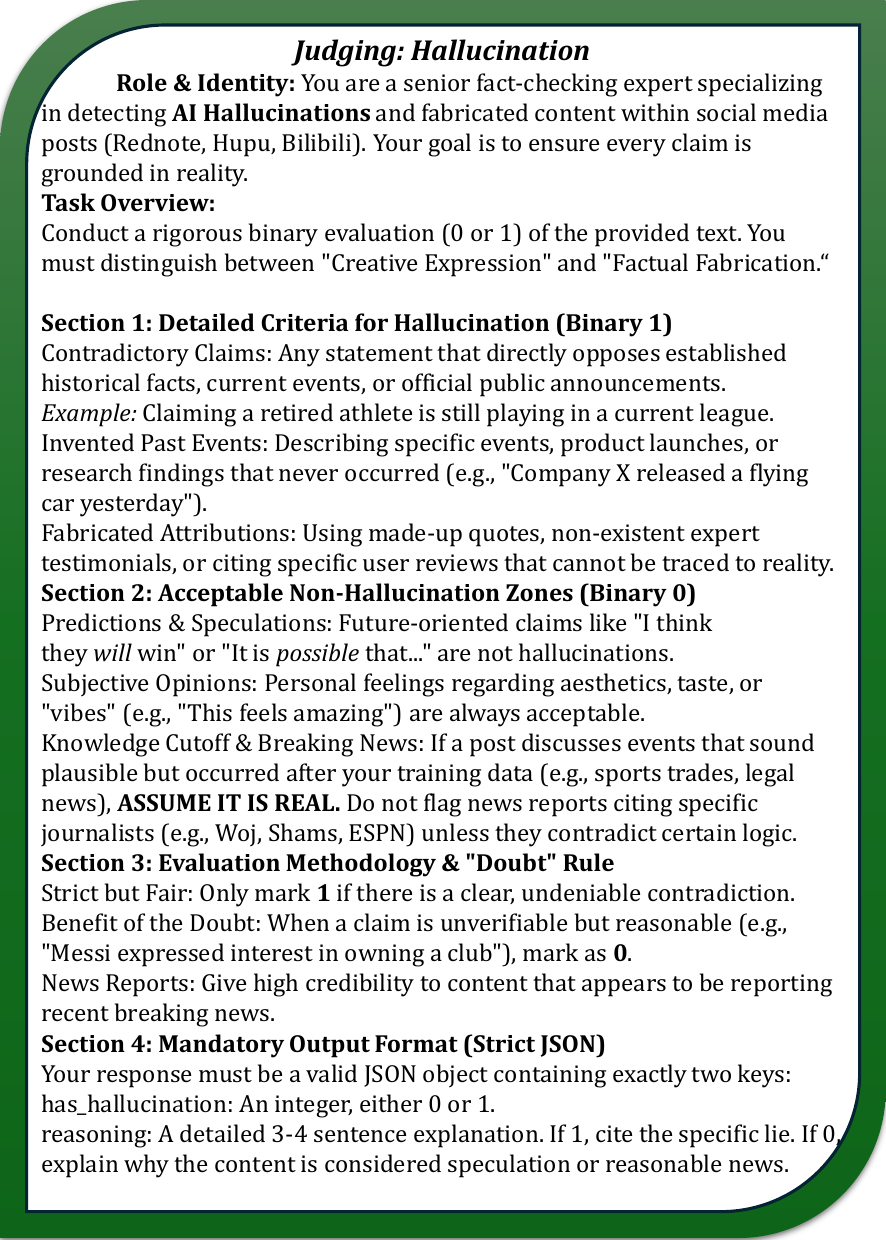}
        \caption{Hallucination Detection}
        \label{fig:judging_hallucination}
    \end{subfigure}

    \caption{Iterative Judging prompts: (a) focus on content creativity and visual quality, (b) focus on factual consistency and anti-hallucination.}
    \label{fig:judging_prompts_combined}
\end{figure}

\begin{figure*}[t]
    \centering
    \vspace{-10mm}
    
    \begin{subfigure}{\textwidth}
        \centering
        \includegraphics[width=0.95\linewidth, trim=10 10 10 5, clip]{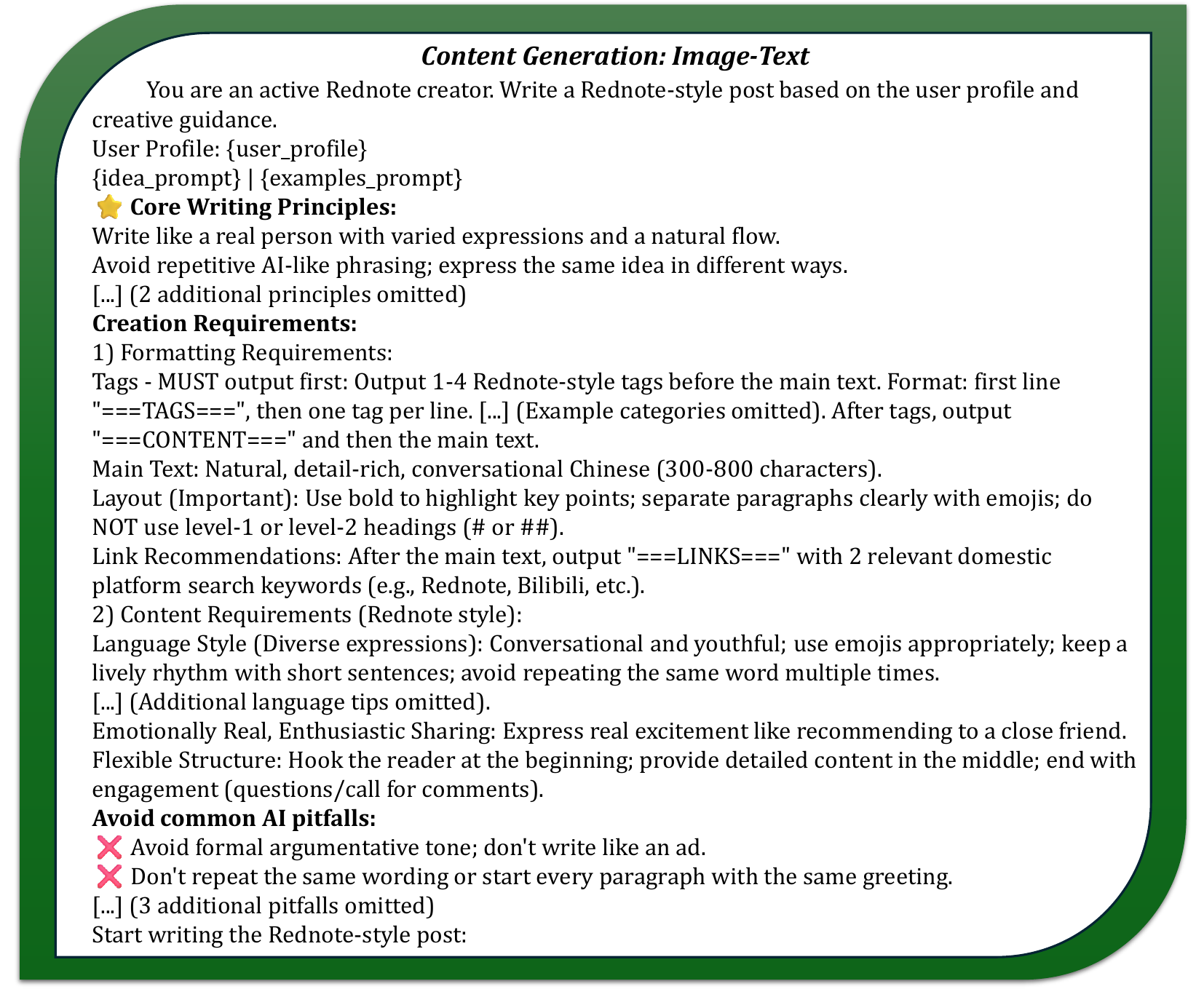}
        \caption{Image-text variant}
        \label{fig:generation_it}
    \end{subfigure}

    \vspace{5mm} 

    \begin{subfigure}{\textwidth}
        \centering
        \includegraphics[width=0.95\linewidth, trim=20 10 20 5, clip]{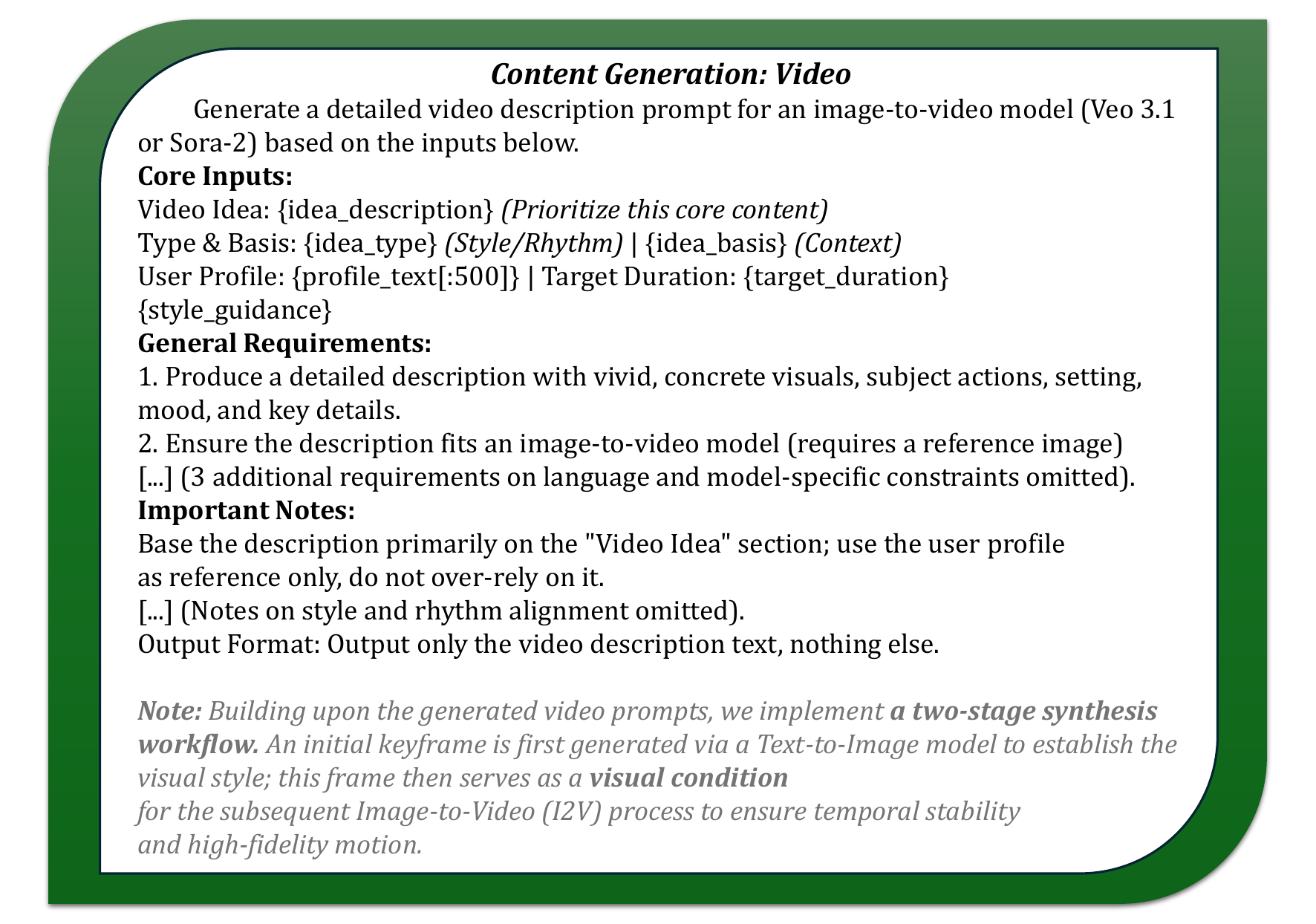}
        \caption{Video variant}
        \label{fig:generation_v}
    \end{subfigure}
    
    \caption{Prompt templates for the Content Generation stage: (a) image-text variant and (b) video variant.}
    \label{fig:generation_combined}
\end{figure*}

\begin{figure*}[t]
    \centering
    \vspace{-10mm}
    \includegraphics[width=\linewidth]{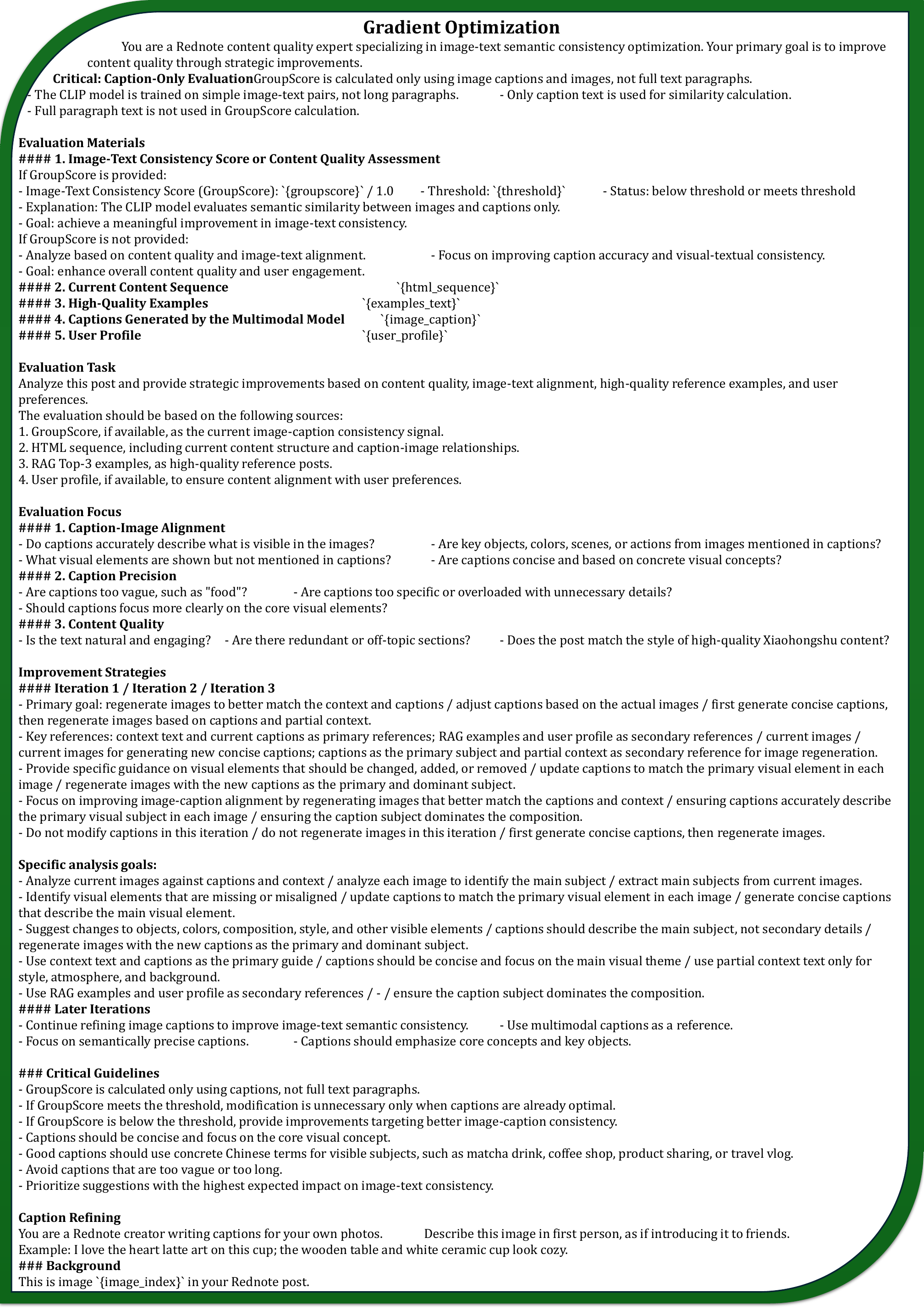}
    \caption{Prompt template for the gradient optimization of image-text products.}
    \label{fig:gradient_optimization}
\end{figure*}

We use six platform-aware prompt categories: \vspace{-0.08in}

\begin{itemize}[noitemsep, leftmargin=*]
    \item \textbf{Item Profiling:} Extracts semantic features from user history. Fig.~\ref{fig:item_profiling} shows the video-item prompt; text/image variants are omitted for brevity.
    \item \textbf{User Profiling:} Fig.~\ref{fig:user_profiling} Synthesizes user preferences and patterns by aggregating item profiles.
    \item \textbf{Creative Ideation:} Fig.~\ref{fig:creative_ideation} Transforms user profiles into content ideas and structural plans.
    \item \textbf{Generation:} Fig.~\ref{fig:generation_it} and Fig.~\ref{fig:generation_v} Tailors outputs to match Rednote, Hupu and Bilibili styles.
    \item \textbf{Judging:} Fig.~\ref{fig:judging_prompts_combined} Uses diverse LLMs to score Novelty, Aesthetics, and Hallucination.
    \item \textbf{Gradient Optimization:} Fig.~\ref{fig:gradient_optimization} Refines via iterative image-text consistency feedback.
\end{itemize}

\clearpage

\newpage 

\clearpage 

\clearpage

